%% file: main.tex
\documentclass{article}

\usepackage{microtype}
\usepackage{graphicx}
\usepackage{booktabs} %

\usepackage{hyperref}

\newif\ifmydraft\mydrafttrue
\newif\ifanonymous\anonymousfalse
\newif\iffull\fulltrue

\ifanonymous
\usepackage{icml2024}
\else
\usepackage[accepted]{icml2024}
\fi

\usepackage{amsmath}
\usepackage{amssymb}
\usepackage{mathtools}
\usepackage{amsthm}

\usepackage[capitalize,noabbrev]{cleveref}

\theoremstyle{plain}
\newtheorem{theorem}{Theorem}[section]

\theoremstyle{definition}
\newtheorem{definition}[theorem]{Definition}

\theoremstyle{remark}

\usepackage[textsize=tiny]{todonotes}

\usepackage{amssymb}
\usepackage{xcolor}
\usepackage{xspace}
\usepackage{enumitem}
\usepackage{subcaption}
\theoremstyle{plain}
\newtheorem{example}{Example}

\begin{document}

\twocolumn[
    \icmltitle{Symbol Correctness in Deep Neural Networks Containing Symbolic Layers}

    \icmlsetsymbol{equal}{*}

    \begin{icmlauthorlist}
        \icmlauthor{Aaron Bembenek}{melb}
        \icmlauthor{Toby Murray}{melb}
    \end{icmlauthorlist}

    \icmlaffiliation{melb}{School of Computing and Information Systems, University of Melbourne, Victoria, Australia}

    \icmlcorrespondingauthor{Aaron Bembenek}{aaron.bembenek@unimelb.edu.au}
    \icmlcorrespondingauthor{Toby Murray}{toby.murray@unimelb.edu.au}

    \icmlkeywords{neurosymbolic AI}

    \vskip 0.3in
]

\printAffiliationsAndNotice{}  %

\input{macros}

\begin{abstract}
    \input{abstract}

\end{abstract}

\input{body}

\bibliography{main}
\bibliographystyle{icml2024}

\newpage
\appendix
\onecolumn %
\input{appendix}

\end{document}

%% file: macros.tex
\newcommand{\SHOWCOMMENT}[1]{\ifmydraft {$\blacksquare$ [{#1}]}\xspace\fi}
\newcommand{\TODO}[1]{\SHOWCOMMENT{\color{red}{#1}}}
\newcommand{\toby}[1]{\SHOWCOMMENT{TM: \color{teal}{#1}}}
\newcommand{\aaron}[1]{\SHOWCOMMENT{AB: \color{blue}{#1}}}
\newcommand{\xxx}{\TODO{XXX}}
\newcommand{\tocite}{\TODO{cite}}

\newcommand{\net}{NS-DNN\xspace}
\newcommand{\nets}{NS-DNNs\xspace}
\newcommand{\definedas}{\ensuremath{\triangleq}}
\newcommand{\many}[1]{\ensuremath{\mathbf{#1}}}
\newcommand{\smooth}[1]{\ensuremath{\tilde{#1}}}
\newcommand{\pseudolabel}{\ensuremath{\psi_x}}

\newlist{questions}{enumerate}{2}
\setlist[questions,1]{label=RQ\arabic*.,ref=RQ\arabic*}
\setlist[questions,2]{label=(\alph*),ref=\thequestionsi(\alph*)}

\newcommand{\shortcite}[1]{\yrcite{#1}}

\newcommand{\autodiff}{\textsc{Multiple}\xspace}
\newcommand{\closest}{\textsc{Closest}\xspace}
\newcommand{\random}{\textsc{Random}\xspace}

%% file: abstract.tex
To handle AI tasks that combine perception and logical reasoning,
recent work introduces Neurosymbolic Deep Neural Networks (\nets), which contain---in addition to traditional neural layers---\emph{symbolic layers}: symbolic expressions (e.g., SAT formulas, logic programs) that are evaluated by solvers during inference.
We identify and formalize an intuitive, high-level principle that can guide the design and analysis of \nets: \emph{symbol correctness}, the correctness of the  intermediate symbols predicted by the neural layers with respect to a (generally unknown) ground-truth symbolic representation of the input data.
We demonstrate that symbol correctness is a necessary property for \net explainability and transfer learning (despite being in general impossible to train for).
Moreover, we show that the framework of symbol correctness provides a precise way to reason and communicate about model behavior at neural-symbolic boundaries, and gives insight into the fundamental tradeoffs faced by \net training algorithms.
In doing so, we both identify significant points of ambiguity in prior work, and provide a framework to support further \net developments.

%% file: body.tex
\section{Introduction}

\begin{figure*}[!t]
    \centering
    \includegraphics[width=0.8\textwidth]{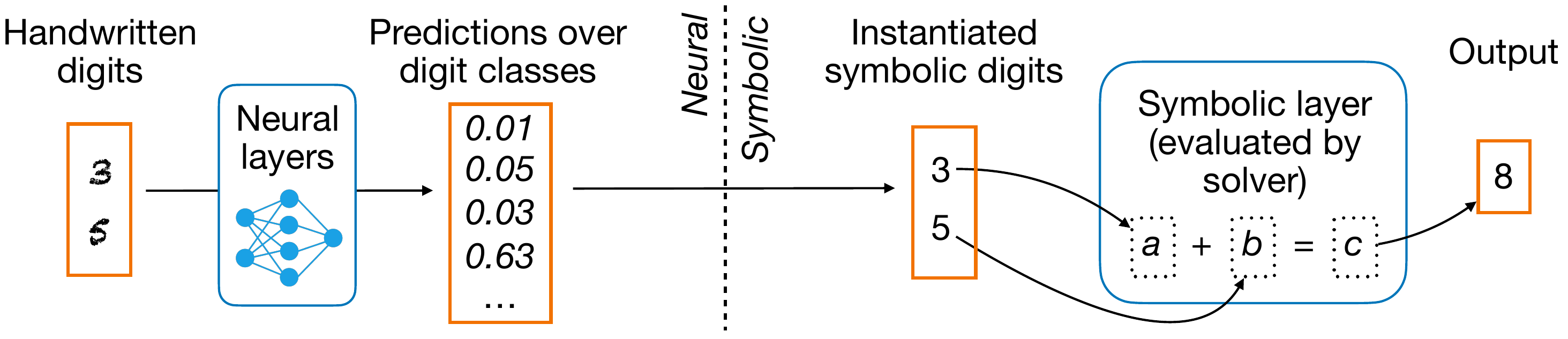}
    \caption{
        This \net for visual addition predicts the sum of two handwritten digits.
        Given images of handwritten digits, neural layers predict the digits' values.
        At the neural-symbolic boundary, these predictions are instantiated as symbolic digits, which are bound to variables $a$ and $b$ in the symbolic layer $a + b = c$.
        A solver is used to evaluate the symbolic layer, assigning the value 8 to the output variable $c$.
        The \net is {\em symbol correct} on the example input: the predicted symbols 3 and 5 match the (unprovided) ground-truth symbols corresponding to the handwritten digits.
        If instead the \net predicted the symbolic digits to be a pair of 4's, it would still be {\em output correct} (since 4 + 4 = 3 + 5), but not symbol correct.
    }
    \label{fig:pipeline}
\end{figure*}

Recent work shows how to embed \emph{symbolic layers}---such as SAT formulas or logic programs---in Deep Neural Networks~\cite{Wang2019SATNet,Vlastelica2020Differentiation,Yang2020NeurASP,Manhaeve2021Neural,Huang2021Scallop,Li2023Scallop,Li2023Softened,Wang2023Grounding}.
The resulting Neurosymbolic Deep Neural Networks (\nets) are more effective than purely neural or purely symbolic approaches on tasks that combine perception with logical or mathematical reasoning.
The canonical \net task is the ``visual addition'' problem~\cite{Manhaeve2021Neural,Yang2020NeurASP,Wang2023Grounding}, where a \net predicts the sum of two handwritten digits (Figure~\ref{fig:pipeline}): first, the neural layers predict the values of the handwritten digits; next, at the neural-symbolic boundary, these predictions are instantiated as symbolic digits bound to variables $a$ and $b$; finally, a solver finds a value for the output variable $c$ in the symbolic expression $a + b = c$.
More complex tasks include solving handwritten Sudoku puzzles~\cite{Yang2020NeurASP,Li2023Softened,Wang2023Grounding,Topan2021Techniques} and performing commonsense reasoning about images~\cite{Yang2020NeurASP,Li2023Scallop}.

When a \net forward pass reaches the boundary between a neural layer and a symbolic layer, the neural layer's real-valued predictions are instantiated as symbols for the subsequent symbolic layer to operate over.
Because training is end-to-end, the neural layers need to learn a mapping from raw input data to these intermediate symbols without any supervision of the symbols.
For example, in visual addition, a pair of handwritten digits is labeled with the mathematical \emph{sum} of the digits, not the individual \emph{summands}.

This paper introduces the concept of \emph{symbol correctness}: correctness at the neural-symbolic boundary with respect to the (unknown) ground-truth symbols.
In the case of visual addition, a symbol-correct \net maps each handwritten digit to its natural symbolic counterpart (e.g., maps a handwritten `3' to the mathematical 3; Figure~\ref{fig:pipeline}).
While this is an intuitive notion, prior work has not clearly and precisely articulated symbol correctness, nor grasped its potential as a powerful principle for thinking about \nets.
We argue that symbol correctness is in fact one of the key concepts for understanding, evaluating, and designing \nets.

After motivating why symbol correctness is such an important concept (Section~\ref{sec:why}), we formalize symbol correctness, contrast it to output correctness (i.e., correctness of the output of the symbolic layer, which is implied by, but does not imply, symbol correctness), and explain why it is generally impossible to train a symbol-correct model without supervision at the neural-symbolic boundary, even with ideal training data and amenable neural components:
symbol correctness requires learning a particular mapping from raw inputs to symbols, but there can exist alternative mappings that are indistinguishable from the desired mapping at the level of the model output (Section~\ref{sec:correctness}).

Building on this negative result, we develop an abstract model of training at the neural-symbolic boundary---in which a training algorithm attempts to learn a symbol-correct representation by integrating neural beliefs and symbolic information about the possible ground-truth symbol---and we place three state-of-the-art training algorithms in the model (Section~\ref{sec:training}).
Having implemented a common framework for training with the three algorithms, we demonstrate that our model helps explain the algorithms' empirical behavior, such as when they struggle to learn symbol-correct representations; doing so clarifies the assumptions underlying the algorithms and their ideal use cases (Section~\ref{sec:eval}).

After reviewing related work (Section~\ref{sec:related_work}), we use the concept of symbol correctness and our abstract model of training to propose future work on \net training (Section~\ref{sec:discussion}).

\section{Symbol Correctness is a Key Concept}\label{sec:why}

We argue for three reasons why symbol correctness is a key concept in the science of \nets.

\subsection{A Desirable Property for \nets}\label{sec:benefits}

Symbol correctness is a desirable property: a symbol-correct \net has benefits over a \net that is not symbol correct, even if both have equally high output accuracy.%
\footnote{In practice, a model would not be symbol correct on all inputs; however, we use ``symbol-correct \net'' to informally refer to a model that is symbol correct on inputs relevant to the exposition.}

A symbol-correct model can provide a satisfactory explanation for a correct prediction, an explanation that is consistent not only with the output, but also with the input data.
In contrast, a model that is not symbol-correct on a point will give an explanation that is consistent with the output, but not with the input (e.g., in the case of visual addition, producing the justification ``4 + 4 = 8'' when the input is actually handwritten `3' and `5').
This, in turn, makes the symbol-correct model more useful in situations where trust is required.

A symbol-correct model is naturally modular, and supports transfer learning at the neural-symbolic boundary.
Say the symbolic layer is the last layer of the \net (as is typical): if the symbolic logic is replaced with new logic, the resulting model will produce the correct \emph{output} on any input the original model is symbol correct on (assuming the new logic is itself correct).
For example, in the case of visual addition, replacing the symbolic layer $a + b = c$ with the symbolic layer $a - b = c$ would give a model for ``visual subtraction,'' without any additional training.

In a symbol-correct model, reasoning about the symbolic layer can be extrapolated to the neural components preceding it.
For example, a simple property one could prove about the symbolic layer $a + b = c$ is $c \ge a \wedge c \ge b$ (assuming $a, b \ge 0$).
For a symbol-correct visual addition model, we can claim that, ``The model output is at least as large as either \emph{handwritten} digit''---that is, our claim can refer to the raw input to the model.
If the model is not symbol correct, our claim is limited to the symbolic layer: ``The model output is at least as large as either \emph{perceived} digit.''

\subsection{Clarity at the Neural-Symbolic Boundary}

Prior work recognizes the importance of what transpires at the neural-symbolic boundary, but the lack of a precise language for talking about correctness at the neural-symbolic boundary has led to ambiguous claims and concepts.
The language of symbol correctness can help on these fronts.

First, by providing a precise notion of correctness at the neural-symbolic boundary, symbol correctness helps evaluate claims about \nets.
For example, Wang et al.~\shortcite{Wang2023Grounding} claim that their \net for visual addition is interpretable, the first step in their argument being that, ``in order to make a correct inference, the neural network must learn to encode MNIST digits in their correct bitvector representation.''
What exactly is the ``correct'' representation?
If we interpret this to mean ``symbol-correct,'' we know to demand a justification for the claim: it suggests that output correctness implies symbol correctness, which---the theory of symbol correctness tells us---does not hold in general (Section~\ref{sec:definition}).

Second, the framework of symbol correctness helps clarify ideas that have appeared in multiple publications, but have not been consistently defined.
Some recent papers on \nets interpret the symbol grounding problem (SGP)~\cite{Harnad1990Symbol} as meaning something along the lines of whether a \net can learn to assign symbols to raw inputs without explicit supervision of this mapping (paraphrasing Chang et al.~\shortcite{Chang2020Assessing}).
However, papers differ in the exact character of this mapping: Li et al.~\shortcite{Li2023Softened} suggest it is \emph{any} ``feasible and generalizable mapping,'' whereas Topan et al.~\shortcite{Topan2021Techniques} suggest it is a \emph{particular} one (without defining the preferred mapping in a general way).
We clarify that these two characterizations of the SGP are in fact different by demonstrating that multiple mappings can be feasible and generalizable.
Moreover, if one equates solving the SGP with learning a symbol-correct representation, no \net training algorithm can hope to solve the SGP in general (Section~\ref{sec:impossible}).

\subsection{A Framework for \net Training Algorithms}

During \net training, loss gradients flow across the neural-symbolic boundary, from the symbolic layer to the neural layer immediately preceding it.
How to construct these gradients is both the principal challenge and principal opportunity in \net design.
As it requires pushing loss information backward through the symbolic layer (which need not be differentiable), there is typically not an obvious way to produce these gradients.
On the other hand, carefully constructed gradients can be used to incorporate symbolic knowledge into the neural learning process; for example, in the case of visual addition, if the training label is zero, the training algorithm can use the symbolically derived information that the (non-negative) digits must also be zero to guide the neural layers toward perceiving them as such.

As we argue extensively later, symbol correctness provides a lens for understanding and comparing existing training algorithms for \nets (Sections~\ref{sec:training} and~\ref{sec:eval}), and also suggests avenues for future exploration (Section~\ref{sec:discussion}).
In short, we hypothesize an ideal training algorithm that knows the symbol-correct labels for the training data, and envision \net training algorithms as attempting to simulate this ideal algorithm by reconciling neural beliefs and symbolic information about the value of a ground-truth symbol.

\section{Formalizing Symbol Correctness}\label{sec:correctness}

We formalize \nets (Section~\ref{sec:neurosymbolic}), define symbol correctness (Section~\ref{sec:definition}), and show why symbol correctness is in general impossible to train for (Section~\ref{sec:impossible}).

\subsection{Formalizing Neurosymbolic-DNNs}\label{sec:neurosymbolic}

We formalize \nets where there is a single symbolic layer at the output end of the model.
This is the only architecture we have seen used in existing case studies~\cite{Wang2019SATNet,Vlastelica2020Differentiation,Yang2020NeurASP,Manhaeve2021Neural,Huang2021Scallop,Li2023Scallop,Li2023Softened,Wang2023Grounding}.
Our formalization and arguments naturally extend to more complex DNN architectures (Appendix~\ref{sec:generalizing}).

Our formalization uses four parameters: the input and output domains $\mathcal{X}$ and $\mathcal{Y}$, respectively, and integers $m, n > 0$, which correspond to the size of the last neural layer and the size of the symbolic input, respectively (Appendix~\ref{sec:variable_length} generalizes this to arbitrary domains).
We treat the neural layers as collectively forming a single neural network.

\begin{definition}[\net]
    An \net is a triple $\langle f_\theta, g, p \rangle$.
    The function $f_\theta : \mathcal{X} \rightarrow \mathbb{R}^m$ is a neural network parameterized by weights $\theta$.
    The grounding function $g : \mathbb{R}^m \rightarrow \{0, 1\}^n$ translates unscaled logits into a symbolic representation.
    The function $p : \{0, 1\}^n \rightarrow \mathcal{Y}$ is a symbolic program (which may contain learnable parameters).
\end{definition}

Inference amounts to computing the composite function $p \circ g \circ f_\theta$: an input value is fed into the neural network, producing logits; the logits are converted into a symbolic input, in the form of binary data; the symbolic program computes on this input, producing the model's final output.

The function $g$ is responsible for translating the real-valued logits produced by the neural network into the binary representation expected by the symbolic program.
In some cases, it might just return the bitstring identifying the class predicted by the neural network (e.g., ``The digit is 2''); in cases where the symbolic program is probabilistic~\cite{Manhaeve2021Neural,Huang2021Scallop}, the bitstring might encode the floating point representation of categorical distributions.

\subsection{When is an \net Correct?}\label{sec:definition}

Our formalization of \nets admits multiple definitions of what it means for a model to be ``correct.''

A standard definition of correctness focuses on the output of the symbolic layer, which is also the observable output of the model.
Suppose we have oracle $o : \mathcal{X} \rightarrow \mathcal{Y}$ that defines the correct symbolic layer output for each model input.
We say that an \net is output correct on an input if it produces an output matching the oracle $o$ on that point:

\begin{definition}[Output correct]
    Let function $o : \mathcal{X} \rightarrow \mathcal{Y}$ be the output oracle for \net $M \definedas \langle f_\theta, g, p \rangle$.
    Model $M$ is \emph{output correct} on point $x \in \mathcal{X}$ iff $(p \circ g \circ f_\theta)(x) = o(x)$.
\end{definition}

While output correctness ensures that the \net produces the correct final answer on a point, we propose that there is another critical dimension of correctness not captured by this output-centric view: \emph{symbol correctness}.

For a given neurosymbolic model $\langle f_\theta, g, p \rangle$, we assume the existence of two implicit entities.
First, we assume the existence of an abstraction function $\mu: \mathcal{X} \rightarrow \mathcal{A}$ that maps a raw input to an element in some mathematical domain $\mathcal{A}$; the value $\mu(x)$ is the mathematical abstraction of the raw data $x$.
Second, we assume the existence of a function $\nu: \mathcal{A} \rightarrow \{0,1\}^n$ that encodes an abstract value as a symbol.
These entities are often intuitive and understood by the model designer.
For example, in the case of visual addition, the domain $\mathcal{A}$ is the set of digit pairs $\{0, \dots, 9\} \times \{0, \dots, 9\}$, the function $\mu$ labels each pair of images of handwritten digits with a pair of mathematical digits, and the function $\nu$ might map a pair of mathematical digits to the concatenation of the 4-bit machine integer encoding of each digit.

We define the \emph{ground-truth symbol function $\alpha: \mathcal{X} \rightarrow \{0,1\}^n$} as the composition $\nu \circ \mu$.
Intuitively, this function provides the missing labels at the neural-symbolic boundary.

Whereas output correctness captures the correctness of the \net in terms of the \emph{output} of the program $p$ and the output oracle $o$, symbol correctness captures the correctness of the model in terms of the \emph{input} to the program $p$ and the ground-truth symbol function $\alpha$:

\begin{definition}[Symbol correct]\label{def:symbol_correct}
    Let function $\alpha : \mathcal{X} \rightarrow \{0, 1\}^n$ be the ground-truth symbol function for \net $M \definedas \langle f_\theta, g, p \rangle$.
    Model $M$ is \emph{symbol correct} on point $x \in \mathcal{X}$ if and only if $(g \circ f_\theta)(x) = \alpha(x)$.
\end{definition}

The symbolic layer $p$ is correct if it takes the ground-truth symbol for an arbitrary point $x \in \mathcal{X}$ to the expected output for that point: $(p \circ \alpha)(x) = o(x)$.
If the symbolic layer $p$ of model $M$ is correct and model $M$ is symbol correct on a point, it is also output correct on that point.
However, output correctness on a point does not in general imply symbol correctness on that point, as there might be multiple inputs that the program $p$ can take to the correct output.

\paragraph{Remarks.}
See Appendix~\ref{sec:remarks} for generalizations of the ground-truth symbol function and symbol correctness.

\paragraph{Notation.}
We use shorthand $o_x \definedas o(x)$ and $\alpha_x \definedas \alpha(x)$.

\subsection{Symbol Correctness Cannot be Trained}\label{sec:impossible}

In general, it is impossible to train a \net to be symbol correct using supervision only on the model output.
More precisely,
there exist symbolic layers $p$ such that no observer can be certain that a model containing the layer $p$ is symbol correct, even if the ground-truth outputs are known.
This result holds regardless of the neural architecture, training data, training algorithm, and output accuracy achieved.
The result assumes that no information is leaked about ground-truth symbols, such as their identities or distribution; this restriction precludes pre-training neural layers on symbols.

While end-to-end training optimizes the model with respect to the model's output, output correctness does not imply symbol correctness.
In fact, for some symbolic layers, it is possible to learn a model that is perfectly output correct, but never symbol correct.
This is the case if the symbolic layer $p$ is a constant function, where output correctness clearly provides no guarantee of symbol correctness; however, it is also the case for some more realistic symbolic layers:

\begin{example}[Xor]\label{ex:xor}
    Consider the \net $\langle f_\theta, g, p \rangle$ where the program $p$ is bit xor: $p(\langle w_0, w_1 \rangle) \definedas w_0 \oplus w_1$.
    Say the neural network $f_\theta$ predicts the opposite bits as the ground-truth symbol function $\alpha$:
    $\forall x \in \mathcal{X}, (g \circ f_\theta)(x) = \neg \alpha_x$ (where the negation operator $\neg$ is vectorized).
    On every point, the model is output correct but not symbol correct.
\end{example}

The ground-truth symbol function $\alpha$ represents a particular (preferred) symbolic interpretation of the input data, but there might be alternative interpretations that, at the level of the model output, are indistinguishable from the preferred interpretation.
Say the \net in Example~\ref{ex:xor} takes as input two handwritten bits, and the function $\alpha$ interprets the handwritten character `0' as the mathematical digit 0 and the character `1' as the mathematical number 1.
There is no reason for the model to favor that ``correct'' interpretation over the opposite interpretation (`0' $\mapsto 1$, `1' $\mapsto 0$): from the model's perspective, they are two arbitrary conventions that fit the data equally well.
Without supervision at the neural-symbolic boundary that informs the model which interpretation is preferable, there is no way to guarantee that the model will learn the symbol-correct representation.

The xor program is an extreme example, and for many symbolic layers of interest symbol correctness is more tightly correlated to output correctness.
This gives some hope for learning symbol-correct \nets via end-to-end training.

\section{Training and Symbol Correctness}\label{sec:training}

We develop a model of \net training and use it to specify an ideal, oracular \net training algorithm that knows the ground-truth symbol at each point (Section~\ref{sec:model}).
We frame real-world \net training as an attempt to simulate this ideal training strategy by reconciling two sources of information about the ground-truth symbol: neural beliefs about the value, and symbolic information about what values might be possible (Section~\ref{sec:two_beliefs}).
We place in the model three state-of-the-art \net training algorithms that embody a range of trust in neural beliefs (Section~\ref{sec:algos}).

\subsection{A Model of \net Training}\label{sec:model}

\begin{figure}[!t]
    \centering
    \includegraphics[width=0.85\columnwidth]{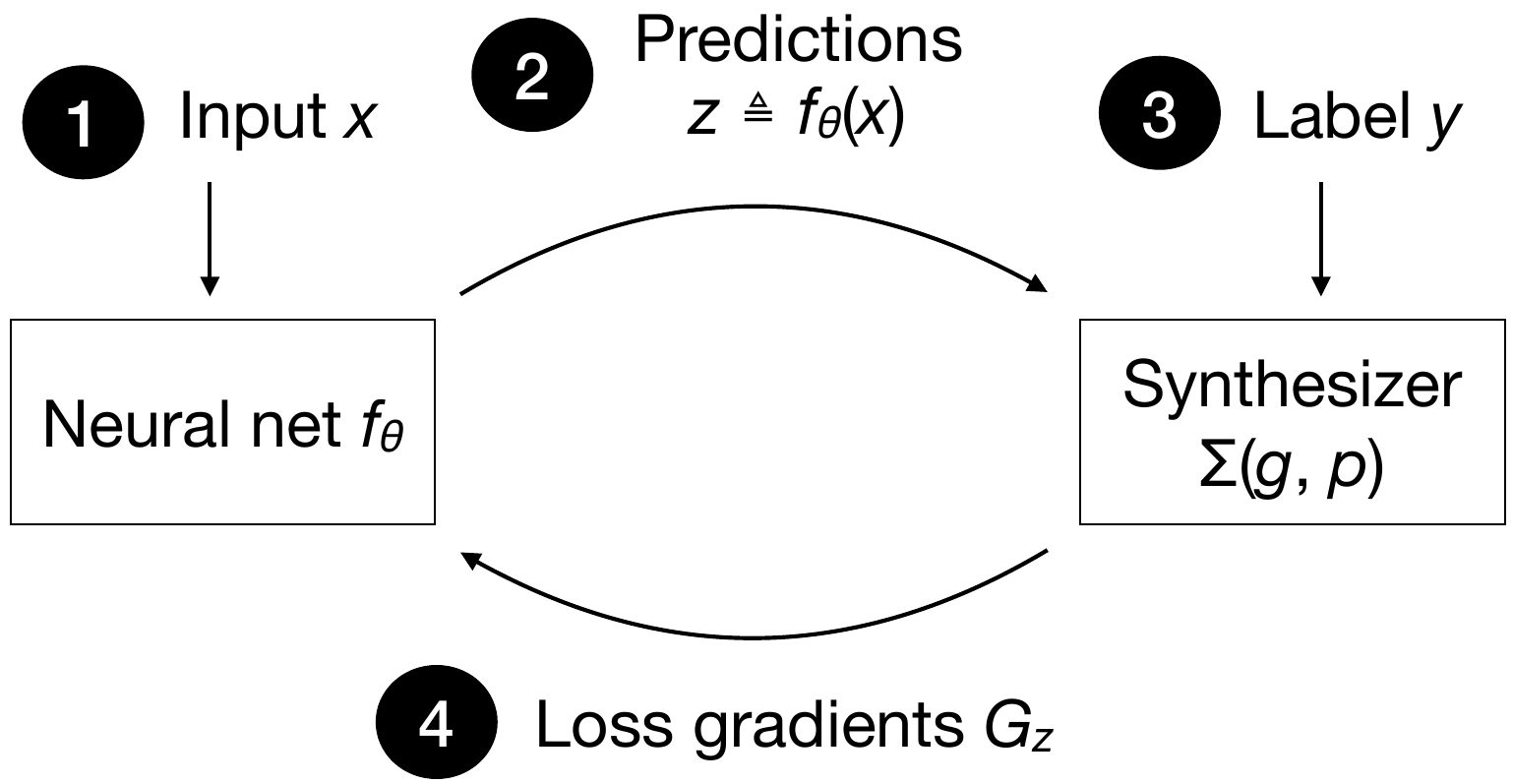}
    \caption{Training on an input $x$, a \emph{synthesizer} $\Sigma(g, p)$ is responsible for building loss gradients $G_z$ from the current neural predictions $z \definedas f_\theta(x)$, training label $y$, grounding function $g$, and symbolic program $p$.}
    \label{fig:training}
\end{figure}

Our model of supervised \net training defines the flow of information that occurs when training a model $\langle f_\theta, g, p \rangle$ on a training sample $\langle x, y \rangle \in \mathcal{X} \times \mathcal{Y}$.
Our assumption of singleton batches is without loss of generality: the high-level flow of information does not depend on the batch size.

The model is based on the premise that the key to understanding \net training is the flow of information into, and between, two components.
The first component is the neural network $f_\theta$.
The second is a \emph{synthesizer} $\Sigma(g, p)$, so called because it synthesizes multiple sources of information to provide feedback to the neural network $f_\theta$.
The design of the synthesizer is what fundamentally differentiates one \net training algorithm from another; we discuss synthesizers after defining the information flows.

Training on a singleton batch $\{\langle x, y \rangle\}$ induces a sequence of four information flows (Figure~\ref{fig:training}):
\begin{enumerate}
    \item Training features $x$ flow into the neural network $f_\theta$.
    \item Neural prediction $z \definedas f_\theta(x)$ flows from the neural network $f_\theta$ into the synthesizer $\Sigma(g, p)$.
    \item Training label $y$ flows into the synthesizer $\Sigma(g, p)$.
    \item Loss gradients $G_z$ flow from the synthesizer $\Sigma(g, p)$ to the neural network $f_\theta$.
\end{enumerate}
After a batch is processed, the neural network updates its parameters using the gradients $G_z$, in turn affecting the neural network's predictions the next time the batch is processed.

The synthesizer $\Sigma(g, p)$ constructs gradients to guide the neural network toward more accurate predictions.%
\footnote{The synthesizer is also responsible for updating any learnable parameters in the program $p$.}
The synthesizer has access to four pieces of information: the grounding function $g$, symbolic program $p$, neural prediction $z \definedas f_\theta(x)$, and the output label $y$.
Different synthesizers use different strategies to synthesize this information into gradients.
The strategies include automatic differentiation and combinatorial constraint solving.

\paragraph{An ideal synthesizer.}
An oracular synthesizer that knows the ground-truth symbol for a point $x$ can return gradients that capture the loss of the current predictions $z$ with respect to the ground-truth value: $G_z \definedas \partial_z \ell(z, \alpha_x)$.%
\footnote{Throughout, metavariable $\ell$ denotes an abstract loss function.}
It provides perfect supervision at the neural-symbolic interface.
As the ground-truth value is generally unknown, a real-world synthesizer has to approximate the ideal gradients.

\subsection{Reasoning with Imperfect Information}\label{sec:two_beliefs}

In our model of \net training, the key challenge facing a synthesizer is to approximate ideal gradients by reconciling two imperfect sources of information about the ground-truth symbolic value: the neural network's belief about the ground-truth value (as encoded in the networks' current predictions $z$), and symbolic information about what values the ground-truth representation could possibly take---that is, the set of symbolic inputs $p^{-1}(\{y\})$ that the program $p$ takes to the label $y$ (implicitly assuming that both the program $p$ and the label $y$ are correct).
Different synthesizers use different strategies to integrate these two sources of information.

If a source of information is known to be highly informative, a simple synthesizer strategy can be effective.
For example, if there is only one bitvector $w_p \in \{0, 1\}^n$ that the program $p$ takes to the label $y$, then the best strategy is to return the gradients $G_z \definedas \partial_z \ell(z, w_p)$: the bitstring $w_p$ must equal the ground-truth symbol $\alpha_x$.
This strategy uses symbolic information exclusively.
On the other hand, if the neural network $f_\theta$ is trustworthy (e.g., has been extensively pre-trained) and predicts a bitstring $w_{z} \in p^{-1}(\{y\})$ with high confidence,
it seems reasonable to trust the neural network and return the gradients $G_z \definedas \partial_z \ell(z, w_z)$.
In this case, the synthesizer uses symbolic information only to verify that the program $p$ takes the bitstring $w_z$ to the label $y$.

More sophisticated synthesizer strategies are necessary in the absence of a highly informative source of information.
Next, we use our model of \net training to analyze the strategies of three state-of-the-art synthesizers.

\subsection{Case Study: Three Synthesizers for Datalog}\label{sec:algos}

We frame three state-of-the-art \net synthesizers in our formal model, analyzing how each synthesizer seeks to emulate the ideal synthesizer despite the absence of the ground-truth symbol $\alpha_x$.
Each synthesizer weights differently the neural network's beliefs about symbol $\alpha_x$; this variance helps explain the synthesizers' empirical behaviors and suggests the proper use for each one (Section~\ref{sec:eval}).

We assume that the symbolic layer $p$ is a Datalog program: a set of logical implications for inferring a database of output facts from a database of input facts~\cite{Ceri1989What}.
The Datalog program $p$ comes equipped with a length-$n$ enumeration of possible input facts and interprets a bitstring $w \in \mathbb{B}^n$ as indicating which of the input facts are enabled (see Appendix~\ref{sec:datalog} for details).
We choose Datalog primarily because it is amenable to many synthesizer techniques.

\subsubsection{Synthesizer \#1: Autodiff}\label{sec:autodiff}

The first synthesizer relaxes the Datalog program $p$ into a smoothed version $\smooth{p}$ that is amenable to automatic differentation~\cite{Baydin2017Automatic}, so that the synthesizer can directly compute the gradients of the loss between the output of program $\smooth{p}$ and the expected output.
The smoothed program $\smooth{p}$ has the same inference rules as the original program $p$, but operates over probabilistic databases of input facts---i.e., each fact is paired with a probability---and in turn produces a probabilistic database of output facts.
In place of the function $g$, the synthesizer uses the function $\smooth{g} : \mathbb{R}^{m} \rightarrow [0, 1]^{n}$ to convert from logits to probabilitistic databases (e.g., by applying the sigmoid and softmax functions as appropriate).

The gradient semi-ring semantics can be used to automatically differentiate the program $\smooth{p}$~\cite{Kimmig2011Algebraic,Manhaeve2021Neural,Huang2021Scallop,Li2023Scallop}.
Given probabilities for the input facts, this semantics computes the probability that each derived fact is true, along with the gradients of that probability with respect to each input fact probability.
The synthesizer uses these gradients to compute the gradient of the loss between the current output and the expected one: $G_z \definedas \partial_z\ell(\smooth{p}(\smooth{g}(z)), y)$.

From the perspective of our model, these gradients reflect multiple weighted guesses about the value of the ground-truth symbol $\alpha_x$.
When there are multiple ways to derive an output fact, the gradients from each derivation are incorporated into the final gradients for that output fact.
For example, in the case of visual addition, if the sum label is 1, the gradients reflect both possible pairs of digits: $\langle 0, 1 \rangle$ and $\langle 1, 0 \rangle$.
Each derivation represents a guess about the value of the ground-truth symbol $\alpha_x$.
As a result of the way that the gradient semi-ring semantics propagates the input fact probabilities---which are given by the neural network---the guesses are weighted so that ones that the neural network predicts to be more likely have a bigger influence on the final gradient.
This reflects the synthesizer's trust in the neural network.
At the same time, the synthesizer hedges its bet by incorporating multiple solutions into the gradients.
For this reason, we refer to it as the ``\autodiff'' synthesizer.

\subsubsection{Synthesizers \#2 \& \#3: Constraint Solving}\label{sec:constraint_based}

As an alternative to automatic differentiation, synthesizers can use combinatorial constraint solving to generate gradients $G_z$; we present two such synthesizers for Datalog.
These algorithms~\cite{Wang2023Grounding,Li2023Softened} were designed for symbolic layers consisting of SMT formulas;
by replacing SMT solving with answer set programming (ASP) solving~\cite{Brewka2011Answer}, the algorithms can be adapted to reason about Datalog~\cite{Bembenek2023SMT}.

Each epoch, each synthesizer produces a pseudolabel $\pseudolabel \in p^{-1}(\{y\})$.
The pseudolabel is the synthesizer's guess for the value of the ground-truth symbol $\alpha_x$: each synthesizer returns gradients $G_z \definedas \partial_z \ell(z, \pseudolabel)$.
Both synthesizers use combinatorial constraint solving to find an appropriate pseudolabel $\pseudolabel$, but vary in how they do so.

\paragraph{Finding the Closest Candidate.}

The first synthesizer, based on SMTLayer~\cite{Wang2023Grounding}, uses discrete optimization to find a bitstring $\pseudolabel$ such that $p(\pseudolabel) = y$ and $\pseudolabel$ is close to the bitstring $g(z)$ currently predicted by the neural network; the latter condition is achieved by weighting choices in the combinatorial encoding proportionally to the predictions $z$ and then finding a maximizing solution.
This strategy is highly trusting of the neural network's predictions: for its guess of the symbol $\alpha_x$, it chooses whichever candidate solution the neural network says is most likely;
unlike automatic differentiation, it does not hedge its bet by incorporating multiple possible solutions into the gradient.
We refer to this synthesizer as ``\closest.''

\paragraph{Sampling a Random Candidate.}

The second synthesizer, based on the soft-grounding strategy~\cite{Li2023Softened}, randomly samples a pseudolabel $\pseudolabel$ from the space $p^{-1}(\{y\})$ by performing a random walk from the pseudolabel chosen in the previous epoch.
During the walk, the synthesizer uses random perturbation and constraint solving to identify a bitstring $w \in p^{-1}(\{y\})$ as a potential next state; it steps to this state if the state is closer to the current neural predictions than the current state (as measured by loss), or by random chance.
At first, these random steps away from the neural predictions occur frequently; an annealing factor makes them less likely as training progresses.
This synthesizer is skeptical of the neural predictions: it does not prioritize solutions that are most in line with the neural predictions, and even sometimes actively chooses a pseudolabel that is in disagreement with the neural predictions.
The insight of Li et al.~\shortcite{Li2023Softened} is that this skepticism should decrease over time, as there becomes (presumably) a better basis for neural beliefs.
We refer to this synthesizer as ``\random.''

\section{Experiments}\label{sec:eval}

We show empirically that, under a reasonable choice of hyperparameters, each synthesizer from Section~\ref{sec:algos}---\autodiff, \closest, and \random---can lead to a visual addition model that is symbol correct on significantly fewer points than it is output correct on;
moreover, the circumstances under which this happens are consistent with our model of \net training.
We focus on visual addition because it is an easily understood example where the synthesizers demonstrate interesting behavior around symbol correctness; our intent differs from that of prior work~\cite{Huang2021Scallop,Li2023Scallop,Manhaeve2021Neural,Li2023Softened,Wang2023Grounding} that demonstrates that these algorithms can solve more complex AI tasks (including ones involving more varied and rich datasets than MNIST digits).

\subsection{Setup}

We built a PyTorch~\cite{Paszke2019PyTorch} framework for Datalog-based \nets with support for all three synthesizers.%
\footnote{\ifanonymous Available in the supplementary material. \else To be made publicly available upon publication. \fi}
For automatic differentiation, the framework uses Scallop~\cite{Huang2021Scallop,Li2023Scallop}; for constraint solving, it uses the ASP solver Clingo~\cite{Gebser2019Multi}.
While we instantiated the framework for visual addition, it is easy to support other tasks.

For each experiment, we used 30k training and 5k test samples, each consisting of two MNIST digits labeled by their mathematical sum, split into batches of size 16.
For the neural layers, we used the architecture from the visual addition network of Li et al.~\shortcite{Li2023Scallop}, and the Adam optimizer with learning rate 0.0001.
For each configuration, we train for 15 epochs and report on the model from the epoch with the highest training accuracy.
We report both output and symbol accuracy: the proportion of test points the model is output correct and symbol correct on, respectively.

\paragraph{Experiment \#1: Varying the Initial Neural Network.}

The goal of this experiment is to see how sensitive each synthesizer is to initial neural network parameters: a synthesizer that is more trusting of neural beliefs (in our model of \net training) should be more sensitive to the initial neural network than a synthesizer that puts more weight on symbolic information.
We ran each synthesizer on 100 randomly initialized neural networks (induced by varying the random seed used by PyTorch), in all cases using the same training and testing data (constructed by randomly sampling MNIST digits without replacement).

\paragraph{Experiment \#2: Skewing the Input Distribution.}\label{sec:exp2}

The goal of this experiment is to see how sensitive each synthesizer is to a mismatch between the actual distribution of inputs and the distribution of inputs assumed by the symbolic logic of the synthesizer.
In particular, whereas the synthesizers assume that every pair of digits is possible, the training and testing data in this experiment consist solely of same-digit pairs: both handwritten digits in an input pair represent the same mathematical digit.
All configurations use the same training and testing data (constructed by sampling MNIST digits with a small amount of replacement).
We ran each synthesizer using 100 different random seeds, but in each case with the same initial neural network: a model from Experiment \#1 with high output and symbol accuracy.
This models using (extremely effective) pre-training.
A synthesizer that puts more weight on symbolic information (in our model of \net training) should perform worse in this setup than a synthesizer more trusting of neural beliefs.

\subsection{Results}

\begin{figure*}[!t]
    \centering
    \begin{subfigure}[t]{0.30\textwidth}
        \includegraphics[width=\textwidth]{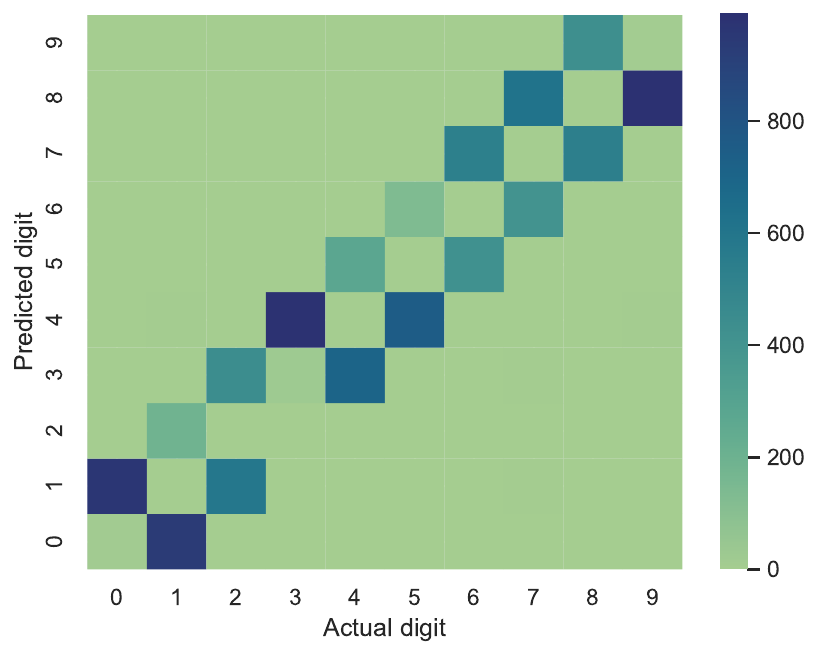}
        \caption{\closest (0.487 vs 0.000)}
        \label{fig:max_sat}
    \end{subfigure}%
    \quad\quad
    \begin{subfigure}[t]{0.30\textwidth}
        \includegraphics[width=\textwidth]{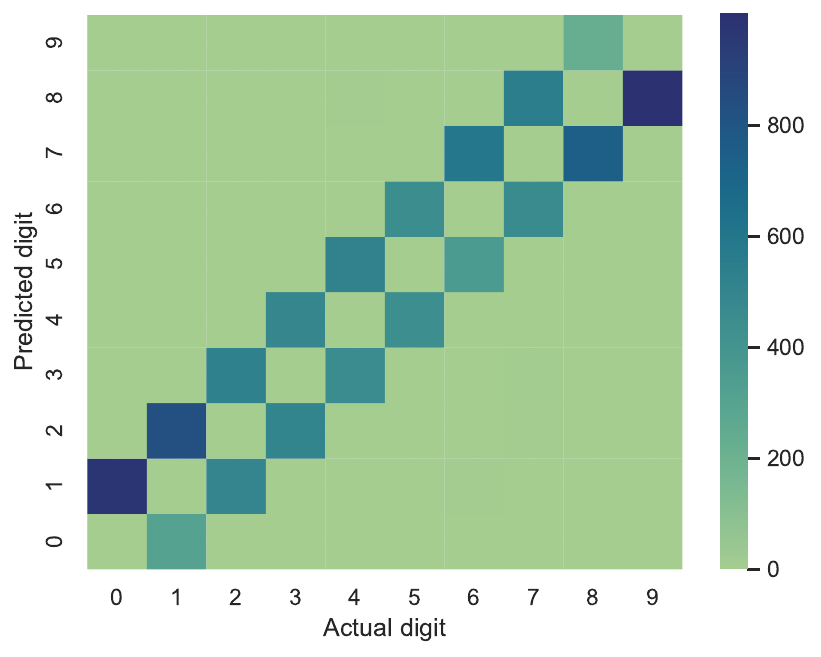}
        \caption{\autodiff (0.489 vs 0.000)}
        \label{fig:autodiff}
    \end{subfigure}%
    \quad\quad
    \begin{subfigure}[t]{0.30\textwidth}
        \includegraphics[width=\textwidth]{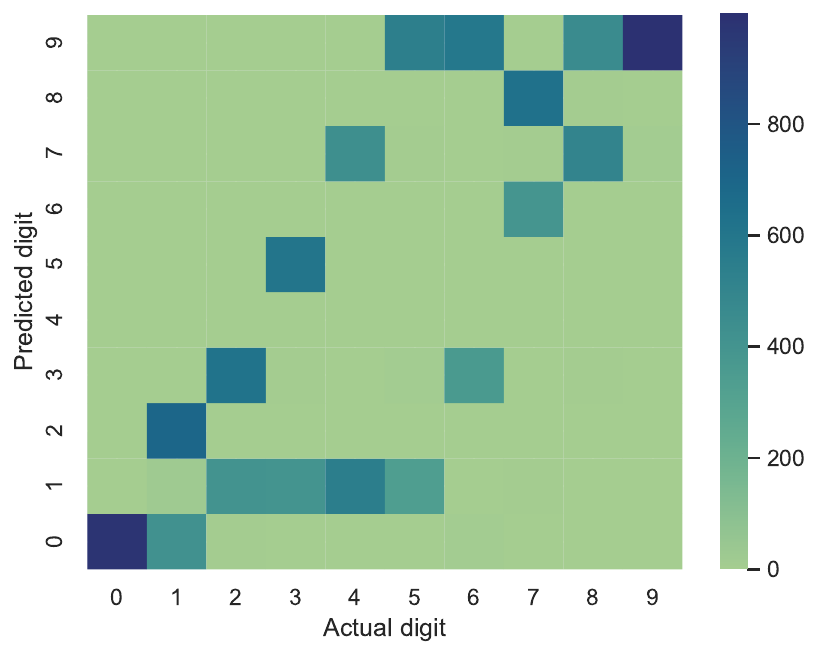}
        \caption{\random (0.581 vs 0.196)}
        \label{fig:soft_ground}
    \end{subfigure}
    \caption{
        Each synthesizer can learn a model with much higher output accuracy than symbol accuracy.
        These heatmaps show the mapping from handwritten digits to mathematical digits in the most extreme model (i.e., having the largest gap between output and symbol accuracy) learned by each synthesizer; we report output accuracy vs symbol accuracy in parentheses.}
    \label{fig:heatmaps}
\end{figure*}

Each synthesizer has cases where it learned a model whose output accuracy is at least 0.3 greater than its symbol accuracy.
For \closest and \autodiff, this happened in 7 and 6 cases, respectively, in Experiment \#1, and never in Experiment \#2;
for \random, this happened in 10 cases in Experiment \#2, and never in Experiment \#1.
These results are consistent with our model of \net training.
(See Appendix~\ref{sec:additional_results} for additional results and discussion.)

On the one hand, our model says that \closest and (to a lesser extent) \autodiff trust neural beliefs about the ground-truth symbol; consequently, it makes sense that they are sensitive to the initial neural network parameters (Experiment \#1).
The fact that they can perform well on skewed data given favorable initial neural network parameters (Experiment \#2) highlights the potential of pre-training for these algorithms, as done in SMTLayer~\cite{Wang2023Grounding}.

On the other hand, our model says that \random is skeptical of neural beliefs, and puts more weight on symbolic information; consequently, it is insensitive to the initial neural network (Experiment \#1), but vulnerable when the distribution of the input data does not match the distribution assumed during symbolic reasoning (Experiment \#2).

The models with much lower symbol accuracy than output accuracy occurred when training led to a local optimum where a digit $d$ is mapped to two digits, $d + k$ and $d - k$ (for $k > 0$); while the model has low symbol accuracy, the sum is consistent with the target on around 50\% of test points (Figure~\ref{fig:heatmaps}).
The details of the learned representations highlight the shortcomings of each synthesizer.
The symbol-incorrect models learned via \closest and \autodiff do not correctly map the digits 0 and 9, despite the fact that symbolic reasoning can label these digits when the target sum is 0 and 18, respectively (as in each case they are the only digits that can lead to the outcome).
This is a missed opportunity: even at its least symbol correct, \random learns a correct mapping for these digits.
Conversely, \random would benefit from taking more heed of statistical patterns that cluster digits: its model confuses digits 5, 6, and 8 with the digit 9, even though 9 is identified clearly.

\section{Related Work}\label{sec:related_work}

Our work on symbol correctness is related to---and often builds on---prior work in neurosymbolic AI.

\subsection{Work Foreshadowing Symbol Correctness}

Our work is the first to formally study symbol correctness and propose it as a central principle for designing and analyzing frameworks for \nets.
Nonetheless, our work builds on and generalizes observations made in prior work.
Yi et al.~\shortcite{Yi2018Neural} briefly note that an interpretable visual-question-answering (VQA) system should produce a correct symbolic representation in addition to a correct final answer, and report how often this happens in their evaluation.
Li et al.~\shortcite{Li2023Softened} frame the traditional and \emph{softened} symbol grounding problems as optimization problems, which they interpret as games between symbolic reasoning and model training.
Compared to such high-level mathematical equations, our lower-level model captures more operational aspects of \net training; additionally, the equations of Li et al.\ do not contain a notion equivalent to symbol correctness.
The theorem of SMTLayer's optimality~\cite{Wang2023Grounding} is conditioned on the symbolic program being injective; our model shows that under this condition it is easy to simulate the ideal synthesizer (provided the program is invertible).
The analyses of Li et al.~\shortcite{Li2023Softened} and Wang et al.~\shortcite{Wang2023Grounding} both include a function with the same type as the function $\alpha$, but do not treat it as providing the ground truth.

\subsection{Other Types of Neurosymbolic Systems}

By combining statistical pattern recognition (via neural networks) and logical reasoning, \nets are one form of neurosymbolic AI~\cite{Garcez2023Neurosymbolic}.
In many cases, \net training can be thought of as neurosymbolic programming~\cite{Chaudhuri2021Neurosymbolic}, where machine learning is used to synthesize programs consisting of neural and symbolic components.
The intuition behind symbol correctness is not limited to DNNs containing symbolic layers: it should apply to neurosymbolic systems more broadly, whenever there is information flowing across a neural-to-symbolic boundary---e.g., VQA systems that are not DNNs~\cite{Yi2018Neural,Mao2019Neuro,Li2020Closed}.

\section{Outlook}\label{sec:discussion}

Symbol correctness is a key concept in the science of \nets.
Our model of \net training---as the balancing of neural beliefs and symbolic information about the unknown value of a ground-truth symbol---suggests three directions for building training regimes around symbol correctness; each one leverages more certain symbolic information to more easily learn a symbol-correct model.

First, curriculum learning regimes might train \nets on data points in order of increasing symbolic uncertainty about the ground-truth symbol, similarly to how some VQA systems train on increasingly complex scenarios~\cite{Mao2019Neuro}.
In visual addition, training on data points labeled with the sum 0 enables learning the correct mapping for the digit 0; subsequently training on data points labeled with the sum 1---while keeping the mapping for 0 mostly fixed---enables learning the correct mapping for the digit 1.

Second, selectively skewing training data can transform \net training problems so that symbolic information is more useful.
For example, if a model for visual addition is trained on same-digit pairs---and the synthesizer is aware of this bias---the symbolic layer can be treated during training as being injective (each sum label has only one possible symbolic input), making it easy to be symbol correct.

Third, the \net research community might study more relaxed training scenarios where a small proportion of symbols are labeled.
Given that symbol correctness implies output correctness (assuming the symbolic layer is correct), future research could explore methodologies for selectively labeling symbols to help synthesizers avoid the type of local minima seen in our experiments (where output accuracy got stuck around 0.5), perhaps in a setting where synthesizers can dynamically request that a symbol be labeled.

\section*{Broader Impact}

By proposing symbol correctness as a fundamental property for understanding, evaluating, and designing \nets,
our work stands to advance the development and use of \nets (and similar neurosymbolic systems).
First, by leveraging the idea of symbol correctness to get to the core of existing \net training algorithms, our model of \net training lowers the barrier to designing new \net training algorithms and highlights natural directions for future research in \net training.
Second, by clarifying the key properties and fundamental mechanisms of \nets, foundational work like ours provides a basis for reasoning---informally and formally---about \nets, a prerequisite for deploying \nets in high-stakes applications.
At a high level, our work shows that neurosymbolic training needs to be done thoughtfully, with a synthesizer strategy that matches the characteristics of the particular learning problem.
As is the nature of foundational work, the societal impact of our work will ultimately depend on the future development and downstream uses of \nets.

\ifanonymous
\else
    \section*{Acknowledgements}
    This work was supported by the joint CATCH MURI-AUSMURI.
    We thank Andrew Cullen and Neil Marchant for their thoughtful feedback on and helpful edits of an earlier version of this paper.
    This research was supported by The University of Melbourne’s Research Computing Services and the Petascale Campus Initiative.
\fi

%% file: appendix.tex
\section{Generalizing to Additional \net Architectures}

\subsection{Beyond Two-Stage Pipelines}\label{sec:generalizing}

\paragraph{NB:} This subsection assumes the reader is familiar with material from Sections~\ref{sec:correctness} and~\ref{sec:training} in the main text.

Our formalization assumes a two-stage \net pipeline: a neural component feeds into a symbolic layer, which produces the model output.
However, the formalization (and our arguments based on it) can be naturally generalized to architectures matching the regular expression $(NS)\!+\!N?$, where $N$ stands for ``\underline{n}eural component,'' and $S$ for ``\underline{s}ymbolic layer.''

First, symbol and output correctness still apply in an $NSN$ pipeline.
The key is that ``output correctness'' refers to the output of the symbolic layer, not the output of the entire model.
Consequently, the correct symbolic output on a training point---$o_x$---no longer corresponds to the label for that point.
During training, the synthesizer would not have access to the sample label $y$; instead, the synthesizer receives gradients backpropagated by the second neural component.
These gradients capture the change of the loss function with respect to the output $(p \circ g)(z)$ produced by the symbolic program, and can be used by the synthesizer to construct an approximation of the value $o_x$ (as suggested in SMTLayer~\cite{Wang2023Grounding}).
The synthesizer in an $NSN$ architecture thus reasons with less certain symbolic information than in an $NS$ architecture.

Second, the function $f_\theta$ in the \net $\langle f_\theta, g, p \rangle$ can itself be an $NSN$ pipeline.
Inductively, the formalization as it stands can thus account for \nets of the form $(NS)\!+$; it would be easy to extend the formalization to also support a terminal neural component.
If there are $k$ symbolic layers $p_1, \dots, p_k$ in the total pipeline (and thus $k$ neural-to-symbolic boundaries), there would be $k$ ground-truth symbol functions $\alpha_1, \dots, \alpha_k$, and thus $k$ different types of symbol correctness for that \net (symbol correctness at the first neural-symbolic boundary, at the second, etc.).

\subsection{Arbitrary Data Domains at the Neural-Symbolic Boundary}\label{sec:variable_length}

Our formalization assumes that the neural component produces predictions of fixed length $m$, and the symbolic layer accepts inputs of fixed length $n$.
The assumption of fixed-size inputs and outputs makes for a more concrete exposition, and also matches the architectures we have seen in prior work~\cite{Wang2019SATNet,Vlastelica2020Differentiation,Yang2020NeurASP,Manhaeve2021Neural,Huang2021Scallop,Li2023Scallop,Li2023Softened,Wang2023Grounding}.
However, the assumption is not fundamental, and our formalization could instead be parameterized by arbitrary domains $\mathcal{U}, \mathcal{V}$, such that $f_\theta : \mathcal{X} \rightarrow \mathcal{U}$, $g : \mathcal{U} \rightarrow \mathcal{V}$, and $p : \mathcal{V} \rightarrow \mathcal{Y}$.
For example, the domain $\mathcal{U}$ could be the set of arbitrary-length vectors of reals, and the domain $\mathcal{V}$ could be the set of arbitrary-length bit vectors.

\section{Additional Remarks on Symbol Correctness}\label{sec:remarks}

\subsection{Partial Function for the Ground-Truth Symbol}

Our formalization defines the ground-truth symbol function $\alpha$ as the composition $\nu \circ \mu$ of two functions, $\mu : \mathcal{X} \rightarrow \mathcal{A}$ and $\nu : \mathcal{A} \rightarrow \{0, 1\}^n$.
The intuition for the function $\mu$ is that it abstracts from raw data to an element in a mathematical domain; in the case of visual addition, it maps from images to a pair of mathematical digits.
Properly, one might define the function $\mu$ as a \emph{partial} (instead of total) function, on the intuition that some raw inputs do not map to an element in the mathematical domain; for example, in visual addition, there is no natural way to map an image of a cat to a pair of digits.
In this case, the ground-truth symbol function $\alpha$ would also be partial, defined on those elements for which the partial function $\mu$ is defined.
As a consequence, there would be no way for a model to be symbol correct on an input that is outside the domain of definition of the partial function $\alpha$ (which intuitively seems appropriate, since that input has no natural correspondence to a symbol).
However, for the sake of simplifying exposition, in the main text we ignore the possibility of out-of-distribution inputs, and treat the function $\alpha$ as total.

\subsection{Relaxed Notions of Symbol Correctness}

Our definition of symbol correctness (Definition~\ref{def:symbol_correct}) requires the neural network $f_\theta$ to exactly predict the ground-truth symbol: $(g \circ f_\theta)(x) = \alpha(x)$.
However, in some situations, exact equality might be too strong.
In particular, if a symbol represents continuous data (such as a probability distribution), then it might be sufficient---i.e., the benefits of symbol correctness might apply (Section~\ref{sec:benefits})---if the predicted symbol is sufficiently close to the ground-truth symbol.
In this case, a suitable relaxation of symbol correctness could be defined, requiring the predicted symbol to be within a certain distance of the ground-truth symbol with respect to an appropriate distance metric.

\section{Datalog as a Symbolic Layer}\label{sec:datalog}

For the symbolic layer, our case study uses Datalog, a simple language for expressing and computing logical inferences~\cite{Ceri1989What}.
A Datalog program is a set of inference rules, where each rule is a Horn clause:
$$\forall \many{x}.[q_0(\many{t}) \leftarrow q_1(\many{t}) \wedge \cdots \wedge q_n(\many{t})].$$
Here, each atom $q_i(\many{t})$ is a relation symbol $q_i$ applied to a vector of terms $\many{t}$, where each term is a number $n$ or a variable $x$ bound by the universal quantifier (\textbf{bold} metavariables denote repetition).
Given a database of input facts---ground atoms in the form $q(\many{n})$---a Datalog program computes the least fixed point of applying the rules in the program to the input facts and any subsequently derived facts; the output of the program is the set of all derived (non-input) facts.

We choose Datalog because it is simple enough to be easily amenable to multiple synthesizer strategies, while also being expressive enough to be interesting: it is used in such varied domains as program analysis~\cite{Whaley2004Cloning,Bravenboer2009Strictly}, networking~\cite{Loo2006Declarative}, and security~\cite{Li2003Datalog,Dougherty2006Specifying}.
We are not aware of other \net developments that use traditional (non-probabilistic) Datalog as a symbolic layer, although Scallop~\cite{Huang2021Scallop} does use a probabilistic version of Datalog.

An \net $\langle f_\theta, g, p \rangle$ with a Datalog symbolic layer has the following characteristics.
The function $p : \{0, 1\}^{|\mathcal{I}|} \rightarrow \{0, 1\}^{|\mathcal{O}|}$ is a Datalog program equipped with enumerations of possible input facts $\mathcal{I}$ and possible output facts $\mathcal{O}$.
The neural network $f_\theta : \mathcal{X} \rightarrow \mathbb{R}^{|\mathcal{I}|}$ predicts the likelihood that each input fact holds; the function $g : \mathbb{R}^{|\mathcal{I}|} \rightarrow \{0,1\}^{|\mathcal{I}|}$ converts those predictions into boolean judgments.
The program $p$ interprets the $i$th bit in its input bitstring as denoting whether the $i$th input fact in the enumeration $\mathcal{I}$ is enabled; the $j$th bit in its output bitstring denotes whether the $j$th output fact in the enumeration $\mathcal{O}$ has been produced.
The ground-truth symbol function $\alpha : \mathcal{X} \rightarrow \{0,1\}^{|\mathcal{I}|}$ maps each raw data point to a database of input Datalog facts.

\begin{example}[Datalog-based visual addition]
    In a Datalog-based \net $\langle f_\theta, g, p \rangle$ for visual addition, the Datalog program $p$ has two input relations ($\mathsf{digit1}$ and $\mathsf{digit2}$), one output relation ($\mathsf{sum}$), and a single inference rule:%
    \footnote{We assume a built-in addition operation (common in Datalog implementations); otherwise, additional inference rules are required.}
    $$\forall x, y.~\mathsf{sum}(x + y) \leftarrow \mathsf{digit1}(x) \wedge \mathsf{digit2}(y).$$
    The length-20 enumeration $\mathcal{I}$ of input facts is $$\langle \mathsf{digit1}({0}), \dots, \mathsf{digit1}(9), \mathsf{digit2}(0), \dots, \mathsf{digit2}(9) \rangle.$$
    The length-19 enumeration $\mathcal{O}$ of output facts is $$\langle \mathsf{sum(0)}, \dots, \mathsf{sum(18)} \rangle.$$
    The neural network $f_\theta$ outputs vectors $z = \langle z_0, \dots, z_{19}\rangle$.
    The function $g$ takes the vector $z$ and returns the concatenation of one-hot encoded vectors $\mathbf{e}_i, \mathbf{e}_j \in \{0, 1\}^{10}$, where $i \in \mathsf{argmax}(\langle z_0, \dots, z_9 \rangle)$ and $j \in \mathsf{argmax}(\langle z_{10}, \dots, z_{19} \rangle)$.
\end{example}

\section{Additional Experimental Results}\label{sec:additional_results}

In this appendix, we give more details on the output and symbol accuracy achieved by each synthesizer during our experiments.
All reported means are arithmetic.
We ran experiments on an HPC cluster with NVIDIA A100 GPUs, varying the amount of memory and number of CPU cores as appropriate for each synthesizer.
We do not focus on training speed; however, it took roughly twice as long to train using the constraint-based synthesizers than with automatic differentiation.
\ifanonymous
    Our experimental data, scripts, and analyses are available in the supplementary material.
\else
    Our experimental data, scripts, and analyses will be made publicly available upon publication.
\fi

\subsection{Experiment \#1: Varying the Initial Neural Network}

\begin{figure}[!t]
    \centering
    \begin{subfigure}[t]{0.47\textwidth}
        \includegraphics[width=\textwidth]{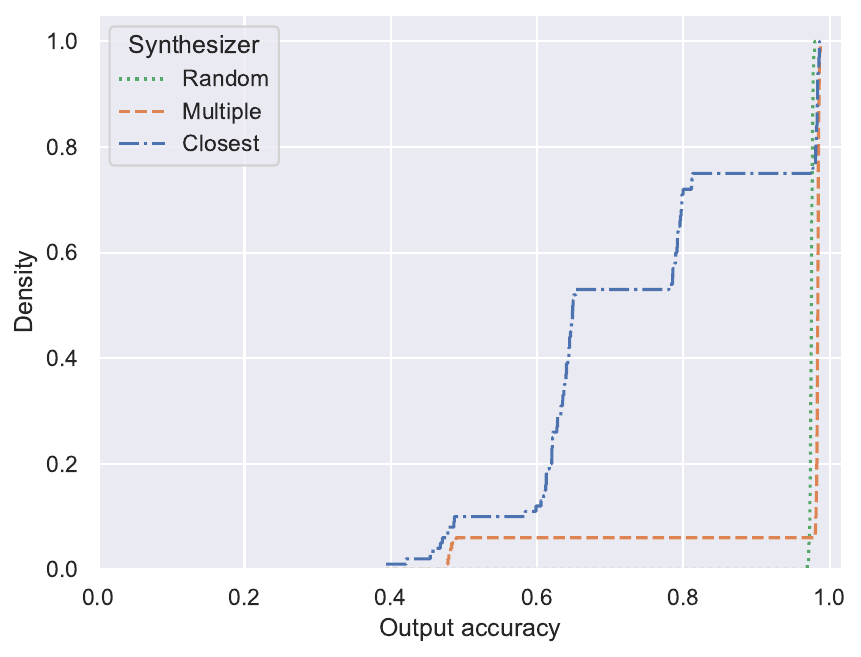}
        \caption{Empirical cumulative density vs output accuracy}
        \label{fig:exp1_output_ecdf}
    \end{subfigure}%
    \quad\quad
    \begin{subfigure}[t]{0.47\textwidth}
        \includegraphics[width=\textwidth]{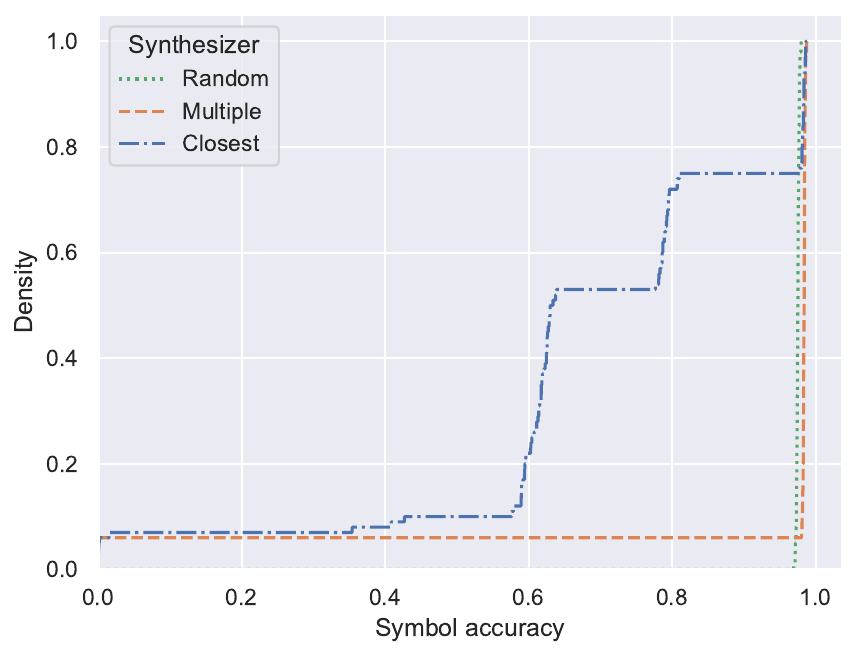}
        \caption{Empirical cumulative density vs symbol accuracy}
        \label{fig:exp1_symbol_ecdf}
    \end{subfigure}
    \caption{
        These empirical cumulative density plots show the proportion of cases for each synthesizer that achieved a given output and symbol accuracy in Experiment \#1.
        \closest demonstrates the highest variability, while \autodiff is bimodal, with a small number of cases of no symbol accuracy (mean: 0.00) and middling output accuracy (mean: 0.48).
        The results for \random appear as a vertical line as all cases achieved roughly the same high output and symbol accuracy (mean: 0.98).
    }
    \label{fig:exp1_ecdf}
\end{figure}

\begin{figure*}[!t]
    \centering
    \begin{subfigure}[t]{0.30\textwidth}
        \includegraphics[width=\textwidth]{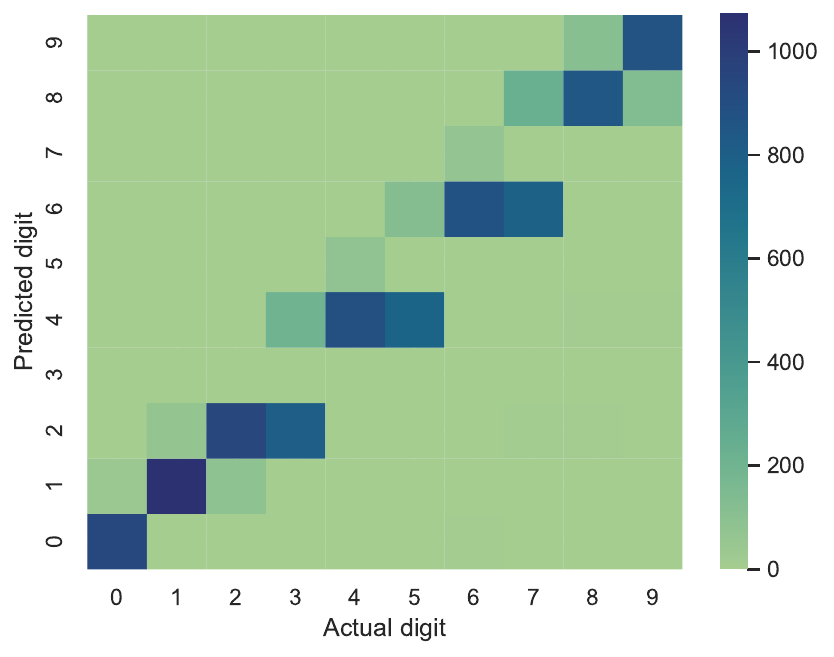}
        \caption{Class B (7 correct digits)}
        \label{fig:max_sat_B}
    \end{subfigure}%
    \quad\quad
    \begin{subfigure}[t]{0.30\textwidth}
        \includegraphics[width=\textwidth]{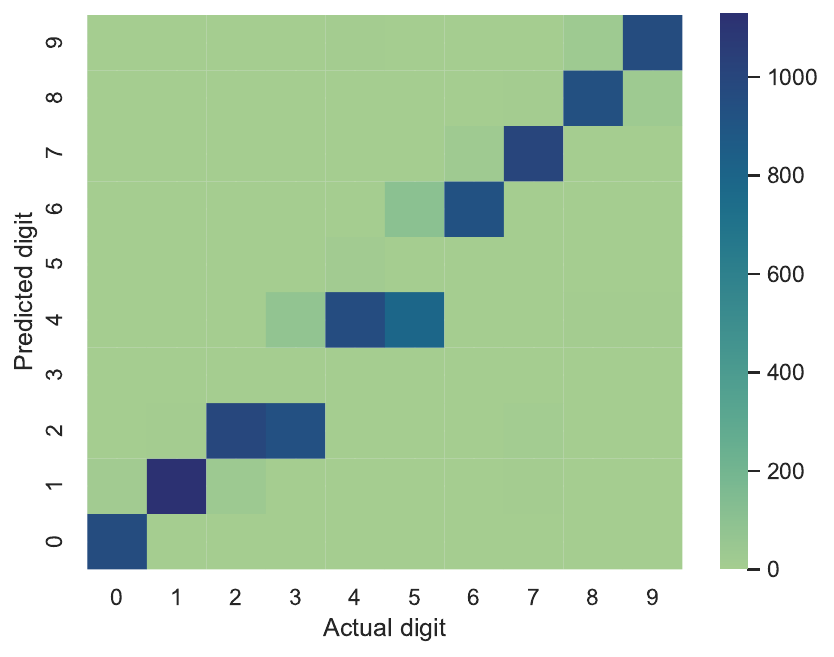}
        \caption{Class C (8 correct digits)}
        \label{fig:max_sat_C}
    \end{subfigure}%
    \quad\quad
    \begin{subfigure}[t]{0.30\textwidth}
        \includegraphics[width=\textwidth]{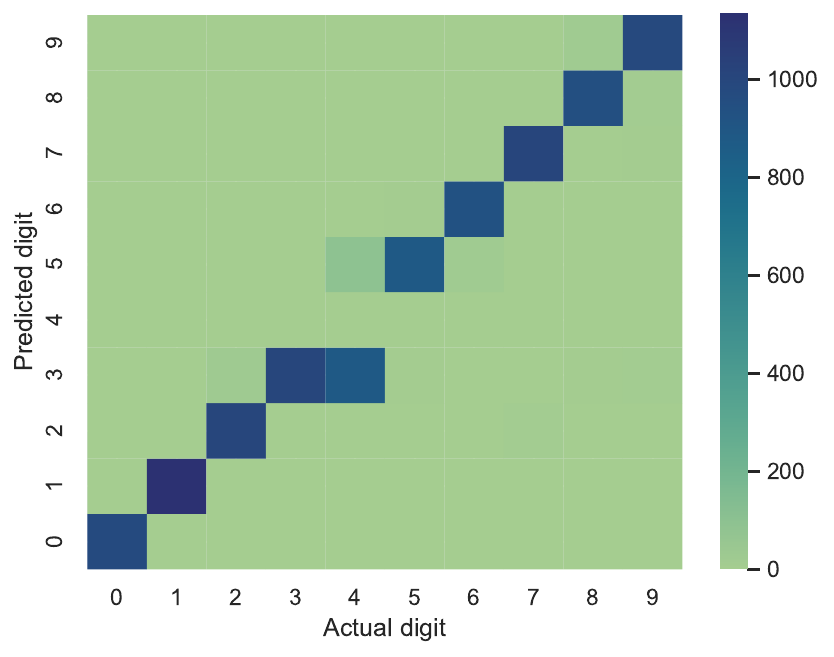}
        \caption{Class D (9 correct digits)}
        \label{fig:max_sat_D}
    \end{subfigure}
    \caption{
        In Experiment \#1, \closest learned models with four varieties of symbol confusion (Table~\ref{tab:max_sat}), three of which are on display here (Class A is displayed in Figure~\ref{fig:max_sat}).
        For the heatmaps, we chose an (arbitrary) representative model for each class.
        Class B models incorrectly map three digits; Class C models, two digits; and Class D models, one digit.
    }
    \label{fig:max_sat_heatmaps}
\end{figure*}

\closest varied greatly with the initial neural network (Figure~\ref{fig:exp1_ecdf}), both in terms of output accuracy (mean: 0.74; stddev: 0.17) and symbol accuracy (mean: 0.70; stddev: 0.25).
The results fall clearly into five classes A-E distinguished by symbol accuracy (Table~\ref{tab:max_sat}; visible as the slopes between plateaus in Figure~\ref{fig:exp1_symbol_ecdf}).
While Class E has high symbol and output accuracy, classes A-D represent four forms of symbol confusion (Figure~\ref{fig:max_sat_heatmaps}).
In Class A, the learned representation is similar to the symbol-incorrect representations learned by automatic differentiation: each digit is translated to its adjacent digits.
In Classes B-D, the learned representation is correct about all digits except three digits, two digits, and one digit, respectively.
This synthesizer gets stuck in local minima because it is highly sensitive to initial neural layer weights: for the vast majority of training samples (mean: 81\%; stddev: 12\%pt.), the pseudolabel never changes from the one given in the first epoch.

\begin{table}[!t]
    \caption{
        The Experiment \#1 results for \closest fall into five categories, based on symbol accuracy (visible as the slopes between plateaus in Figure~\ref{fig:exp1_symbol_ecdf}), and corresponding to the number of correctly mapped digits.
    }
    \label{tab:max_sat}
    \centering
    \begin{small}
        \begin{tabular}{lrrrr}
            \toprule
                  &       & Number of     &                                            \\
                  &       & digits mapped & \multicolumn{2}{c}{Mean accuracy}          \\
            Class & Count & correctly     & Symbol                            & Output \\
            \midrule
            A     & 7     & 0             & 0.00                              & 0.46   \\
            B     & 3     & 7             & 0.40                              & 0.45   \\
            C     & 43    & 8             & 0.61                              & 0.63   \\
            D     & 22    & 9             & 0.79                              & 0.79   \\
            E     & 25    & 10            & 0.98                              & 0.98   \\
            \bottomrule
        \end{tabular}
    \end{small}
\end{table}

\autodiff was bimodal:
In 94 out of 100 cases, it had high output and symbol accuracy (mean: 0.98).
In six cases though, it had relatively low output accuracy (mean: 0.48), and no symbol accuracy (mean: 0.00).

\random achieved consistently good results: mean output and symbol accuracy of 0.98, and trending upwards when our experiments ended after 15 training epochs.
\random takes longer to converge to a high-accuracy model than the other synthesizers as it aggressively adopts new pseudolabels for the first 10 epochs, before iteratively lowering the probability of a random step following the exponential annealing strategy (our implementation follows the original one of Li et al.~\shortcite{Li2023Softened}).

\subsection{Experiment \#2: Skewing the Input Distribution}

\begin{figure}[!b]
    \centering
    \begin{subfigure}[t]{0.47\textwidth}
        \includegraphics[width=\textwidth]{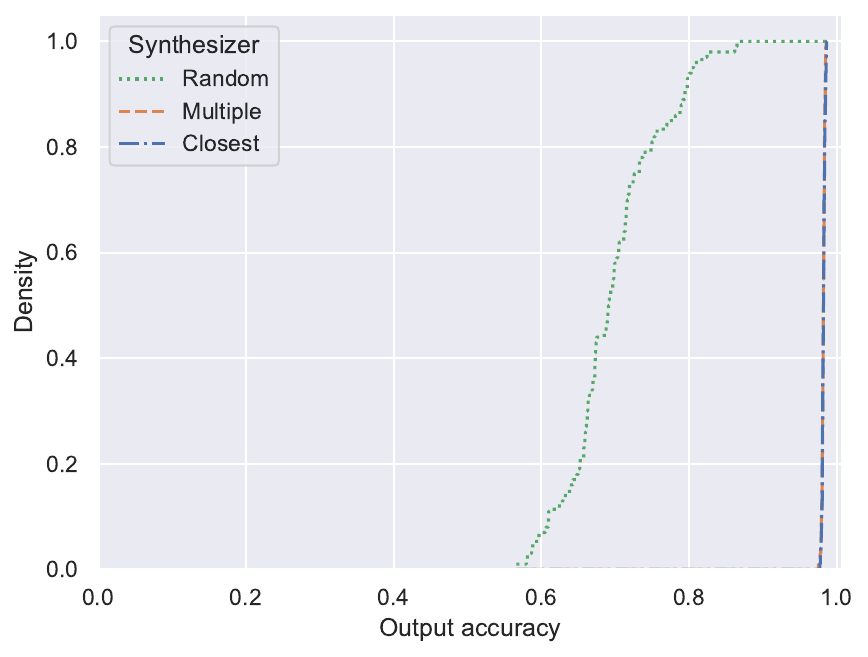}
        \caption{Empirical cumulative density vs output accuracy}
        \label{fig:exp2_output_ecdf}
    \end{subfigure}%
    \quad\quad
    \begin{subfigure}[t]{0.47\textwidth}
        \includegraphics[width=\textwidth]{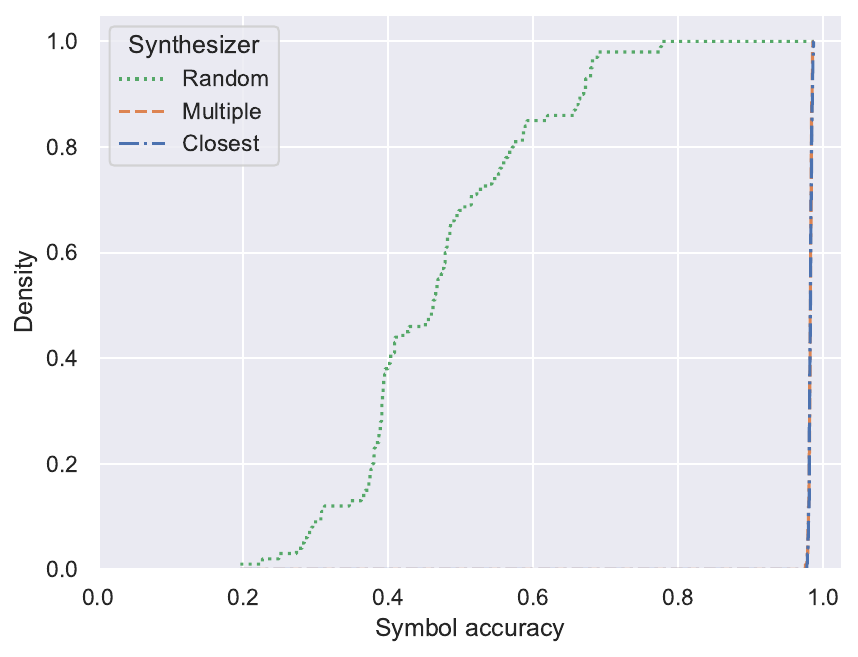}
        \caption{Empirical cumulative density vs symbol accuracy}
        \label{fig:exp2_symbol_ecdf}
    \end{subfigure}
    \caption{
        These empirical cumulative density plots show the proportion of cases for each synthesizer that achieved a given output and symbol accuracy in Experiment \#2.
        \random demonstrates a good deal of variability, and achieved relatively low output and symbol accuracy (maxes of 0.87 and 0.78, respectively).
        The results for \closest and \autodiff appear as (overlapping) vertical lines as all cases achieved roughly the same high output and symbol accuracy (mean: 0.98).
    }
    \label{fig:exp2_ecdf}
\end{figure}

\begin{figure*}[!t]
    \centering
    \begin{subfigure}[t]{0.30\textwidth}
        \includegraphics[width=\textwidth]{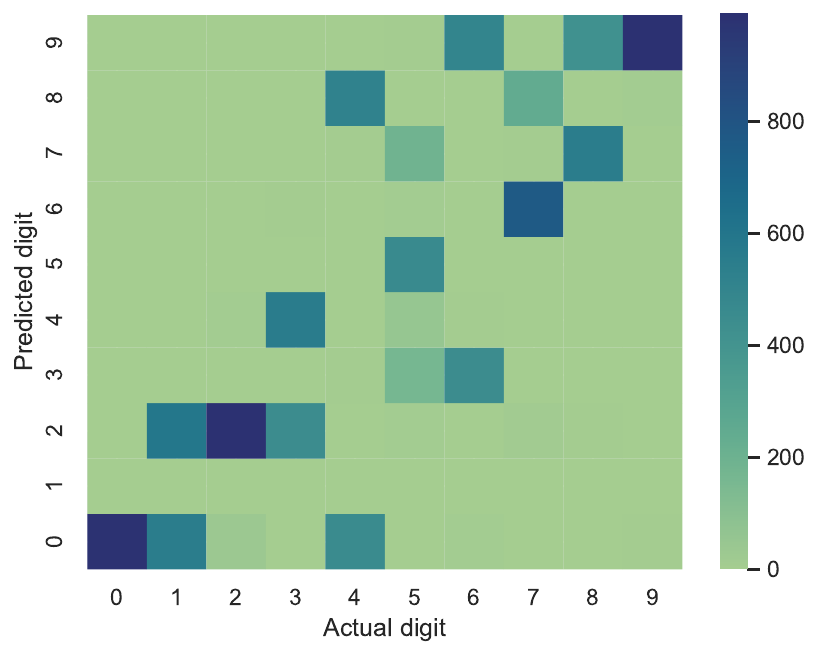}
        \caption{Class B (3 correct digits)}
        \label{fig:soft_ground_B}
    \end{subfigure}%
    \quad\quad
    \begin{subfigure}[t]{0.30\textwidth}
        \includegraphics[width=\textwidth]{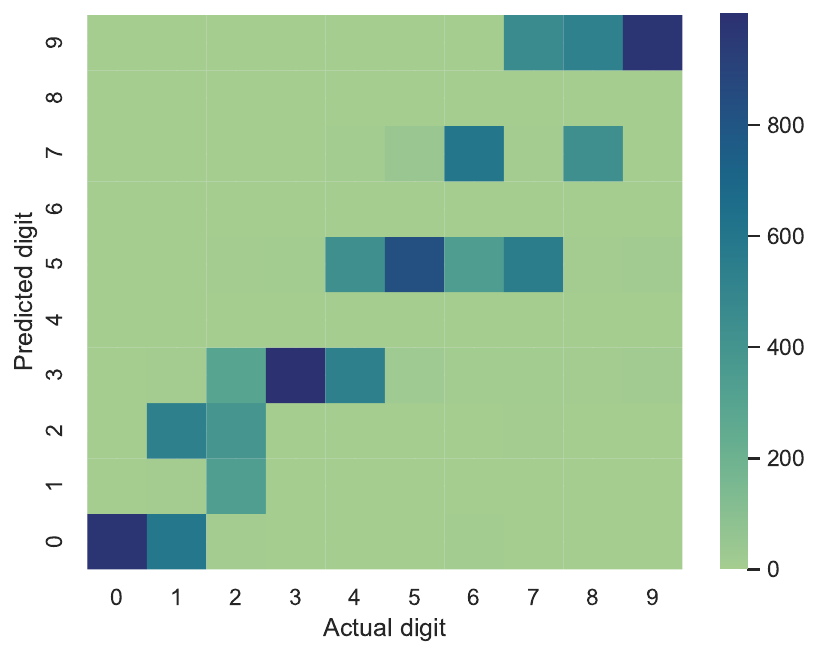}
        \caption{Class C (4 correct digits)}
        \label{fig:soft_ground_C}
    \end{subfigure}%
    \quad\quad
    \begin{subfigure}[t]{0.30\textwidth}
        \includegraphics[width=\textwidth]{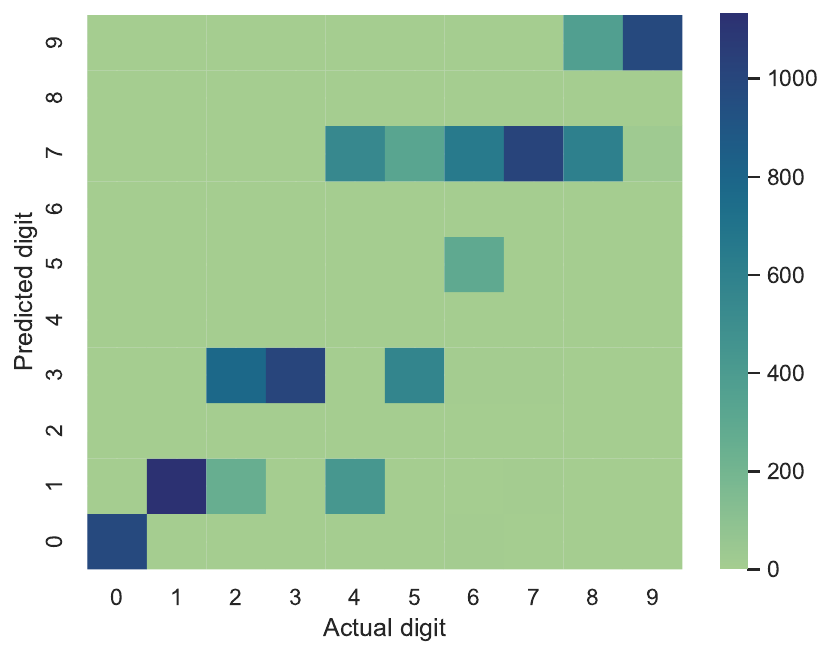}
        \caption{Class D (5 correct digits)}
        \label{fig:soft_ground_D}
    \end{subfigure}
    \begin{subfigure}[t]{0.30\textwidth}
        \includegraphics[width=\textwidth]{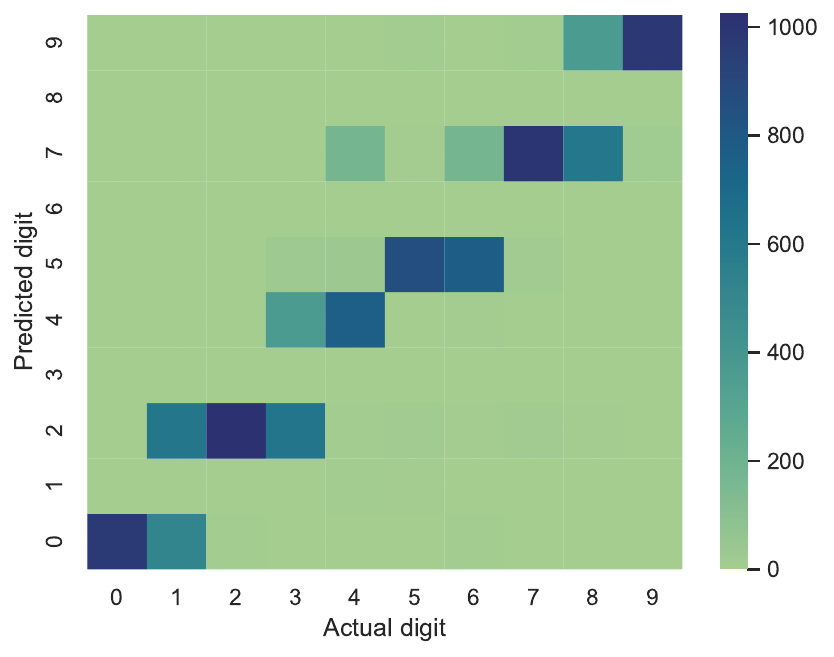}
        \caption{Class E (6 correct digits)}
        \label{fig:soft_ground_E}
    \end{subfigure}%
    \quad\quad
    \begin{subfigure}[t]{0.30\textwidth}
        \includegraphics[width=\textwidth]{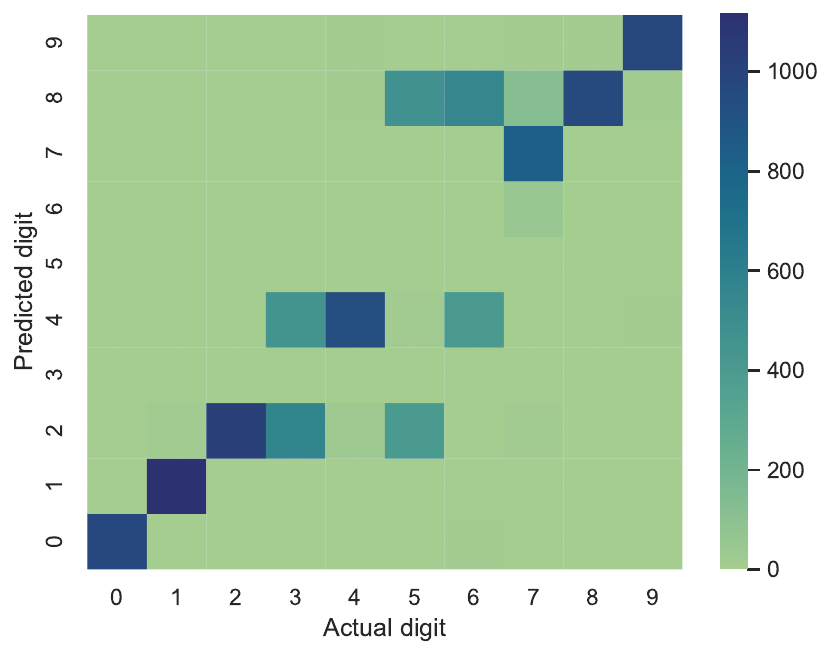}
        \caption{Class F (7 correct digits)}
        \label{fig:soft_ground_F}
    \end{subfigure}%
    \quad\quad
    \begin{subfigure}[t]{0.30\textwidth}
        \includegraphics[width=\textwidth]{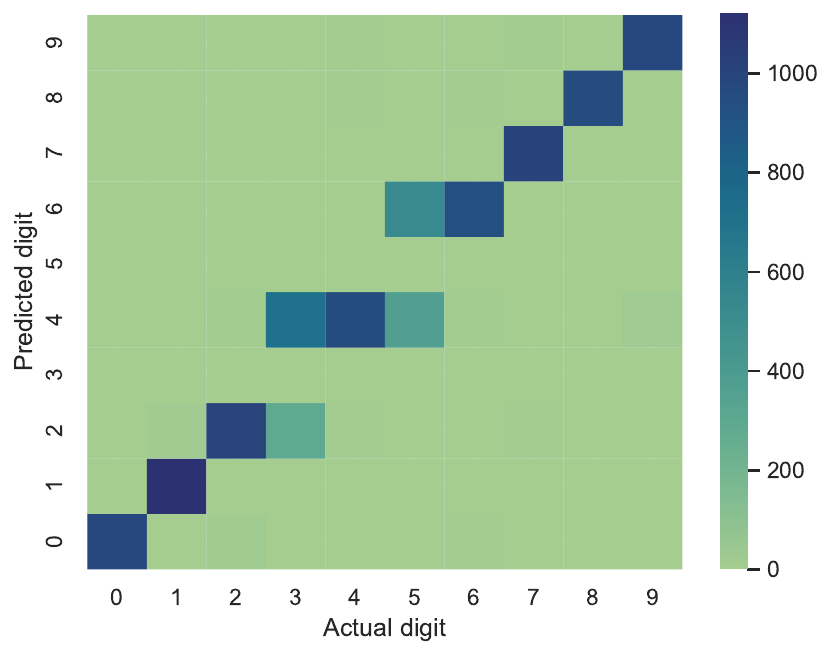}
        \caption{Class G (8 correct digits)}
        \label{fig:soft_ground_G}
    \end{subfigure}
    \caption{
        In Experiment \#2, \random learned models with seven varieties of symbol confusion (Table~\ref{tab:max_sat}), six of which are on display here (Class A is displayed in Figure~\ref{fig:soft_ground}).
        For the heatmaps, we chose an (arbitrary) representative model for each class.
        Class B models correctly map three digits; Class C models, four digits; Class D models, five digits; and so on.
    }
    \label{fig:soft_ground_heatmaps}
\end{figure*}

In Experiment \#2, \closest and \autodiff achieved consistently high output and symbol accuracy (mean: 0.98, with low variance; Figure~\ref{fig:exp2_ecdf}).
\random had worse---and more variable---performance, both in terms of output accuracy (mean: 0.70; stddev: 0.06) and symbol accuracy (mean: 0.47; stddev: 0.12).
The results for \random fall into seven different categories A-G based on symbol accuracy (Table~\ref{tab:soft_ground}; visible as the slopes between plateaus in Figure~\ref{fig:exp2_symbol_ecdf}).
While these classes are not as cleanly delineated as the classes for \closest in Experiment \#1, they do represent qualitatively distinct representations learned by \random (Figure~\ref{fig:soft_ground_heatmaps}).
In particular, Class A models correctly map two digits; Class B models, three digits; Class C models, four digits; and so on.
The digits 0 and 9 are always correctly mapped, reflecting the weight that \random puts on symbolic information (these digits can be unambiguously learned from inputs labeled with the sums 0 and 18, respectively).

\begin{table}[!b]
    \caption{
        The Experiment \#2 results for \random fall into seven categories, based on symbol accuracy (visible as the slopes between plateaus in Figure~\ref{fig:exp2_symbol_ecdf}), and corresponding to the number of correctly mapped digits.
    }
    \label{tab:soft_ground}
    \centering
    \begin{small}
        \begin{tabular}{lrrrr}
            \toprule
                  &       & Number of     &                                            \\
                  &       & digits mapped & \multicolumn{2}{c}{Mean accuracy}          \\
            Class & Count & correctly     & Symbol                            & Output \\
            \midrule
            A     & 3     & 2             & 0.22                              & 0.58   \\
            B     & 10    & 3             & 0.30                              & 0.60   \\
            C     & 31    & 4             & 0.39                              & 0.66   \\
            D     & 27    & 5             & 0.47                              & 0.70   \\
            E     & 14    & 6             & 0.56                              & 0.74   \\
            F     & 13    & 7             & 0.67                              & 0.80   \\
            G     & 2     & 8             & 0.78                              & 0.86   \\
            \bottomrule
        \end{tabular}
    \end{small}
\end{table}

%% file: main.bbl
\begin{thebibliography}{28}
\providecommand{\natexlab}[1]{#1}
\providecommand{\url}[1]{\texttt{#1}}
\expandafter\ifx\csname urlstyle\endcsname\relax
  \providecommand{\doi}[1]{doi: #1}\else
  \providecommand{\doi}{doi: \begingroup \urlstyle{rm}\Url}\fi

\bibitem[Baydin et~al.(2017)Baydin, Pearlmutter, Radul, and Siskind]{Baydin2017Automatic}
Baydin, A.~G., Pearlmutter, B.~A., Radul, A.~A., and Siskind, J.~M.
\newblock Automatic {D}ifferentiation in {M}achine {L}earning: a {S}urvey.
\newblock \emph{Journal of Machine Learning Research}, 18\penalty0 (153):\penalty0 1--43, 2017.

\bibitem[Bembenek et~al.(2023)Bembenek, Greenberg, and Chong]{Bembenek2023SMT}
Bembenek, A., Greenberg, M., and Chong, S.
\newblock From {SMT} to {ASP}: Solver-{B}ased {A}pproaches to {S}olving {D}atalog {S}ynthesis-as-{R}ule-{S}election {P}roblems.
\newblock \emph{Proceedings of the ACM on Programming Languages}, 7\penalty0 ({POPL}):\penalty0 185--217, 2023.
\newblock \doi{10.1145/3571200}.

\bibitem[Bravenboer \& Smaragdakis(2009)Bravenboer and Smaragdakis]{Bravenboer2009Strictly}
Bravenboer, M. and Smaragdakis, Y.
\newblock Strictly {D}eclarative {S}pecification of {S}ophisticated {P}oints-to {A}nalyses.
\newblock In \emph{ACM SIGPLAN Conference on Object-Oriented Programming, Systems, Languages, and Applications}, pp.\  243--262, 2009.
\newblock \doi{10.1145/1640089.1640108}.

\bibitem[Brewka et~al.(2011)Brewka, Eiter, and Truszczynski]{Brewka2011Answer}
Brewka, G., Eiter, T., and Truszczynski, M.
\newblock Answer {S}et {P}rogramming at a {G}lance.
\newblock \emph{Communications of the ACM}, 54\penalty0 (12):\penalty0 92--103, 2011.
\newblock \doi{10.1145/2043174.2043195}.

\bibitem[Ceri et~al.(1989)Ceri, Gottlob, and Tanca]{Ceri1989What}
Ceri, S., Gottlob, G., and Tanca, L.
\newblock What {Y}ou {A}lways {W}anted to {K}now {A}bout {D}atalog ({A}nd {N}ever {D}ared to {A}sk).
\newblock \emph{{IEEE} Transactions on Knowledge and Data Engineering}, 1\penalty0 (1):\penalty0 146--166, 1989.
\newblock \doi{10.1109/69.43410}.

\bibitem[Chang et~al.(2020)Chang, Flokas, Lipson, and Spranger]{Chang2020Assessing}
Chang, O., Flokas, L., Lipson, H., and Spranger, M.
\newblock Assessing {SATNet}'s {A}bility to {S}olve the {S}ymbol {G}rounding {P}roblem.
\newblock In \emph{Advances in Neural Information Processing Systems}, 2020.

\bibitem[Chaudhuri et~al.(2021)Chaudhuri, Ellis, Polozov, Singh, Solar-Lezama, and Yue]{Chaudhuri2021Neurosymbolic}
Chaudhuri, S., Ellis, K., Polozov, O., Singh, R., Solar-Lezama, A., and Yue, Y.
\newblock Neurosymbolic {P}rogramming.
\newblock \emph{Foundations and Trends{\textregistered} in Programming Languages}, 7\penalty0 (3):\penalty0 1--86, 2021.
\newblock \doi{10.1561/2500000049}.

\bibitem[d'Avila Garcez \& Lamb(2023)d'Avila Garcez and Lamb]{Garcez2023Neurosymbolic}
d'Avila Garcez, A. and Lamb, L.~C.
\newblock Neurosymbolic {AI:} the 3rd wave.
\newblock \emph{Artificial Intelligence Review}, 56\penalty0 (11):\penalty0 12387--12406, 2023.
\newblock \doi{10.1007/S10462-023-10448-W}.

\bibitem[Dougherty et~al.(2006)Dougherty, Fisler, and Krishnamurthi]{Dougherty2006Specifying}
Dougherty, D.~J., Fisler, K., and Krishnamurthi, S.
\newblock Specifying and {R}easoning {A}bout {D}ynamic {A}ccess-{C}ontrol {P}olicies.
\newblock In \emph{International Joint Conference on Automated Reasoning}, pp.\  632--646, 2006.
\newblock \doi{10.1007/11814771_51}.

\bibitem[Gebser et~al.(2019)Gebser, Kaminski, Kaufmann, and Schaub]{Gebser2019Multi}
Gebser, M., Kaminski, R., Kaufmann, B., and Schaub, T.
\newblock Multi-shot {ASP} solving with {C}lingo.
\newblock \emph{Theory and Practice of Logic Programming}, 19\penalty0 (1):\penalty0 27--82, 2019.
\newblock \doi{10.1017/S1471068418000054}.

\bibitem[Harnad(1990)]{Harnad1990Symbol}
Harnad, S.
\newblock The symbol grounding problem.
\newblock \emph{Physica D: Nonlinear Phenomena}, 42\penalty0 (1–3):\penalty0 335--346, 1990.
\newblock \doi{10.1016/0167-2789(90)90087-6}.

\bibitem[Huang et~al.(2021)Huang, Li, Chen, Samel, Naik, Song, and Si]{Huang2021Scallop}
Huang, J., Li, Z., Chen, B., Samel, K., Naik, M., Song, L., and Si, X.
\newblock Scallop: {F}rom {P}robabilistic {D}eductive {D}atabases to {S}calable {D}ifferentiable {R}easoning.
\newblock In \emph{Advances in Neural Information Processing Systems}, pp.\  25134--25145, 2021.

\bibitem[Kimmig et~al.(2011)Kimmig, {Van den Broeck}, and {De Raedt}]{Kimmig2011Algebraic}
Kimmig, A., {Van den Broeck}, G., and {De Raedt}, L.
\newblock An {A}lgebraic {P}rolog for {R}easoning about {P}ossible {W}orlds.
\newblock In \emph{{AAAI} Conference on Artificial Intelligence}, pp.\  209--214, 2011.

\bibitem[Li \& Mitchell(2003)Li and Mitchell]{Li2003Datalog}
Li, N. and Mitchell, J.~C.
\newblock Datalog with {C}onstraints: {A} {F}oundation for {T}rust {M}anagement {L}anguages.
\newblock In \emph{International Symposium on Practical Aspects of Declarative Languages}, pp.\  58--73, 2003.
\newblock \doi{10.1007/3-540-36388-2_6}.

\bibitem[Li et~al.(2020)Li, Huang, Hong, Chen, Wu, and Zhu]{Li2020Closed}
Li, Q., Huang, S., Hong, Y., Chen, Y., Wu, Y.~N., and Zhu, S.
\newblock Closed {L}oop {N}eural-{S}ymbolic {L}earning via {I}ntegrating {N}eural {P}erception, {G}rammar {P}arsing, and {S}ymbolic {R}easoning.
\newblock In \emph{International Conference on Machine Learning}, pp.\  5884--5894, 2020.

\bibitem[Li et~al.(2023{\natexlab{a}})Li, Huang, and Naik]{Li2023Scallop}
Li, Z., Huang, J., and Naik, M.
\newblock Scallop: A {L}anguage for {N}eurosymbolic {P}rogramming.
\newblock \emph{Proceedings of the {ACM} on Programming Languages}, 7\penalty0 ({PLDI}):\penalty0 1463--1487, 2023{\natexlab{a}}.
\newblock \doi{10.1145/3591280}.

\bibitem[Li et~al.(2023{\natexlab{b}})Li, Yao, Chen, Xu, Cao, Ma, and L{\"{u}}]{Li2023Softened}
Li, Z., Yao, Y., Chen, T., Xu, J., Cao, C., Ma, X., and L{\"{u}}, J.
\newblock Softened {S}ymbol {G}rounding for {N}euro-symbolic {S}ystems.
\newblock In \emph{International Conference on Learning Representations}, 2023{\natexlab{b}}.

\bibitem[Loo et~al.(2006)Loo, Condie, Garofalakis, Gay, Hellerstein, Maniatis, Ramakrishnan, Roscoe, and Stoica]{Loo2006Declarative}
Loo, B.~T., Condie, T., Garofalakis, M.~N., Gay, D.~E., Hellerstein, J.~M., Maniatis, P., Ramakrishnan, R., Roscoe, T., and Stoica, I.
\newblock Declarative {N}etworking: {L}anguage, {E}xecution and {O}ptimization.
\newblock In \emph{{ACM} {SIGMOD} International Conference on Management of Data}, pp.\  97--108, 2006.
\newblock \doi{10.1145/1142473.1142485}.

\bibitem[Manhaeve et~al.(2021)Manhaeve, Duman{\v{c}}i{\'{c}}, Kimmig, Demeester, and Raedt]{Manhaeve2021Neural}
Manhaeve, R., Duman{\v{c}}i{\'{c}}, S., Kimmig, A., Demeester, T., and Raedt, L.~D.
\newblock Neural probabilistic logic programming in {DeepProbLog}.
\newblock \emph{Artificial Intelligence}, 298:\penalty0 103504, 2021.
\newblock \doi{10.1016/j.artint.2021.103504}.

\bibitem[Mao et~al.(2019)Mao, Gan, Kohli, Tenenbaum, and Wu]{Mao2019Neuro}
Mao, J., Gan, C., Kohli, P., Tenenbaum, J.~B., and Wu, J.
\newblock The {N}euro-{S}ymbolic {C}oncept {L}earner: Interpreting {S}cenes, {W}ords, and {S}entences {F}rom {N}atural {S}upervision.
\newblock In \emph{International Conference on Learning Representations}, 2019.

\bibitem[Paszke et~al.(2019)Paszke, Gross, Massa, Lerer, Bradbury, Chanan, Killeen, Lin, Gimelshein, Antiga, Desmaison, K{\"{o}}pf, Yang, DeVito, Raison, Tejani, Chilamkurthy, Steiner, Fang, Bai, and Chintala]{Paszke2019PyTorch}
Paszke, A., Gross, S., Massa, F., Lerer, A., Bradbury, J., Chanan, G., Killeen, T., Lin, Z., Gimelshein, N., Antiga, L., Desmaison, A., K{\"{o}}pf, A., Yang, E.~Z., DeVito, Z., Raison, M., Tejani, A., Chilamkurthy, S., Steiner, B., Fang, L., Bai, J., and Chintala, S.
\newblock {PyTorch}: An {I}mperative {S}tyle, {H}igh-{P}erformance {D}eep {L}earning {L}ibrary.
\newblock In \emph{Advances in Neural Information Processing Systems}, pp.\  8024--8035, 2019.

\bibitem[Topan et~al.(2021)Topan, Rolnick, and Si]{Topan2021Techniques}
Topan, S., Rolnick, D., and Si, X.
\newblock Techniques for {S}ymbol {G}rounding with {SATNet}.
\newblock In \emph{Advances in Neural Information Processing Systems}, pp.\  20733--20744, 2021.

\bibitem[Vlastelica et~al.(2020)Vlastelica, Paulus, Musil, Martius, and Rol{\'i}nek]{Vlastelica2020Differentiation}
Vlastelica, M., Paulus, A., Musil, V., Martius, G., and Rol{\'i}nek, M.
\newblock Differentiation of {B}lackbox {C}ombinatorial {S}olvers.
\newblock In \emph{International Conference on Learning Representations}, 2020.

\bibitem[Wang et~al.(2019)Wang, Donti, Wilder, and Kolter]{Wang2019SATNet}
Wang, P., Donti, P.~L., Wilder, B., and Kolter, J.~Z.
\newblock {SATN}et: Bridging deep learning and logical reasoning using a differentiable satisfiability solver.
\newblock In \emph{International Conference on Machine Learning}, pp.\  6545--6554, 2019.

\bibitem[Wang et~al.(2023)Wang, Vijayakumar, Lu, Ganesh, Jha, and Fredrikson]{Wang2023Grounding}
Wang, Z., Vijayakumar, S., Lu, K., Ganesh, V., Jha, S., and Fredrikson, M.
\newblock Grounding {N}eural {I}nference with {S}atisfiability {M}odulo {T}heories.
\newblock In \emph{Advances in Neural Information Processing Systems}, 2023.

\bibitem[Whaley \& Lam(2004)Whaley and Lam]{Whaley2004Cloning}
Whaley, J. and Lam, M.~S.
\newblock Cloning-{B}ased {C}ontext-{S}ensitive {P}ointer {A}lias {A}nalysis {U}sing {B}inary {D}ecision {D}iagrams.
\newblock In \emph{ACM SIGPLAN Conference on Programming Language Design and Implementation}, pp.\  131--144, 2004.
\newblock \doi{10.1145/996841.996859}.

\bibitem[Yang et~al.(2020)Yang, Ishay, and Lee]{Yang2020NeurASP}
Yang, Z., Ishay, A., and Lee, J.
\newblock Neur{ASP}: Embracing {N}eural {N}etworks into {A}nswer {S}et {P}rogramming.
\newblock In \emph{International Joint Conference on Artificial Intelligence}, pp.\  1755--1762, 2020.
\newblock \doi{10.24963/IJCAI.2020/243}.

\bibitem[Yi et~al.(2018)Yi, Wu, Gan, Torralba, Kohli, and Tenenbaum]{Yi2018Neural}
Yi, K., Wu, J., Gan, C., Torralba, A., Kohli, P., and Tenenbaum, J.
\newblock Neural-{S}ymbolic {VQA}: Disentangling {R}easoning from {V}ision and {L}anguage {U}nderstanding.
\newblock In \emph{Advances in Neural Information Processing Systems}, pp.\  1039--1050, 2018.

\end{thebibliography}
